\documentclass{article}

\usepackage{microtype}
\usepackage{graphicx}
\usepackage{subfigure}
\usepackage{booktabs} % for professional tables
\usepackage{hyperref}

\usepackage{multirow}
\usepackage{amsmath}
\usepackage{amssymb}
\usepackage{mathtools}
\usepackage{amsthm}
\usepackage[capitalize,noabbrev]{cleveref}
\usepackage[utf8]{inputenc} % allow utf-8 input
\usepackage[T1]{fontenc}    % use 8-bit T1 fonts
\usepackage{booktabs}       % professional-quality tables
\usepackage{nicefrac}       % compact symbols for 1/2, etc.
\usepackage{xcolor}         % colors
\usepackage{latexsym}
\usepackage{multirow}
\usepackage{makecell}
\usepackage{enumitem}
\usepackage{graphicx}
\usepackage{setspace}
\usepackage{subfigure}
\usepackage{latexsym}
\usepackage{inconsolata}
\usepackage{bm}
\usepackage{wrapfig}
\usepackage{arydshln}
\usepackage{hyperref}
\usepackage{url}
\usepackage{pifont}
\usepackage{bbding}

\theoremstyle{plain}

\theoremstyle{definition}

\theoremstyle{remark}

\usepackage[textsize=tiny]{todonotes}

\usepackage[accepted]{icml2024}

\newcommand{\yansen}[1]{{\bf \color{blue} [[Yansen says ``#1'']]}}

\newcommand{\cmark}{\ding{51}}%
\newcommand{\xmark}{\ding{55}}%

\ifdefined\nocomment
\renewcommand{\yansen}[1]{\iffalse{#1}\fi}
\fi
\begin{document}

\twocolumn[

\icmltitle{Efficient and Effective Time-Series Forecasting with Spiking Neural Networks}


\begin{icmlauthorlist}
\icmlauthor{Changze Lv}{fudan}
\icmlauthor{Yansen Wang}{msra}
\icmlauthor{Dongqi Han}{msra}
\icmlauthor{Xiaoqing Zheng}{fudan}
\icmlauthor{Xuanjing Huang}{fudan}
\icmlauthor{Dongsheng Li}{msra}

\end{icmlauthorlist}

\icmlaffiliation{fudan}{School of Computer Science, Fudan University, Shanghai, China}
\icmlaffiliation{msra}{Microsoft Research Asia, Shanghai, China}

\icmlcorrespondingauthor{Xiaoqing Zheng}{zhengxq@fudan.edu.cn}
\icmlcorrespondingauthor{Yansen Wang}{yansenwang@microsoft.com}

\icmlkeywords{Spiking Neural Networks, Time-Series Forecasting}
\vskip 0.3in
]

\printAffiliationsAndNotice{The work was conducted during the internship of Changze Lv (czlv22@m.fudan.edu.cn) at Microsoft Research Asia.}

\begin{abstract}
Spiking neural networks (SNNs), inspired by the spiking behavior of biological neurons, provide a unique pathway for capturing the intricacies of temporal data.
However, applying SNNs to time-series forecasting is challenging due to difficulties in effective temporal alignment, complexities in encoding processes, and the absence of standardized guidelines for model selection.
In this paper, we propose a framework for SNNs in time-series forecasting tasks, leveraging the efficiency of spiking neurons in processing temporal information.
Through a series of experiments, we demonstrate that our proposed SNN-based approaches achieve comparable or superior results to traditional time-series forecasting methods on diverse benchmarks with much less energy consumption.
Furthermore, we conduct detailed analysis experiments to assess the SNN's capacity to capture temporal dependencies within time-series data, offering valuable insights into its nuanced strengths and effectiveness in modeling the intricate dynamics of temporal data.
Our study contributes to the expanding field of SNNs and offers a promising alternative for time-series forecasting tasks, presenting a pathway for the development of more biologically inspired and temporally aware forecasting models.
Our code is available at \url{https://github.com/microsoft/SeqSNN}.
\end{abstract}

\section{Introduction}
\label{sec:introduction}
Spiking neural network (SNN) is regarded as the third generation of neural network \citep{Maas1997NetworksOS} for its energy efficiency, event-driven paradigm, and biological plausibility.
Nowadays, SNNs have achieved comparable performance with artificial neural networks (ANNs) in image classification \citep{hu2018spiking,Fang2021DeepRL,Ding2021OptimalAC,Zhou2022SpikformerWS,yao2023spike}, text classification \citep{lv2023spiking,Lv2023SpikeBERTAL}, sequential image classification \citep{jeffares2022spikeinspired,chen2023unleashing}, and time-series classification \citep{dominguez2018deep, fang2020multivariate}.
% The pretraining schema has come into reality for SNN \citep{zhu2023spikegpt,Lv2023SpikeBERTAL,li2023spikeclip} and achieved remarkable performance on traditional ANN benchmarks, 
%Notably, Spikformer-v2 proposed by \citet{zhou2024spikformer} has achieved comparable performance with traditional ANN model Swin-Transformer \citep{liu2021swin} on ImageNet \citep{Deng2009ImageNetAL}, specifically following a phase of image pretraining.
% However, these works either completely neglect the temporal nature of SNN or bluntly fit the data into the event-driven paradigm despite the sequential data format, e.g., by repeating the samples along the time axis or only preserving changes across data points, which can not fully exploit the advantages of SNNs in the domain of temporal signal processing. \yansen{need to polish}
Nevertheless, the existing studies either completely neglect the temporal nature of SNN or oversimplify the incorporation of data into the event-driven paradigm despite the sequential data format, e.g., by repeating the samples along the time axis \citep{Fang2021DeepRL,lv2023spiking} or only preserving changes across data points \citep{reid2014financial, fang2020multivariate}.
These strategies, while serving their purpose, can not fully exploit the advantages of SNNs in the domain of temporal signal processing.

% \yansen{Maybe it's better to directly phrase this paragraph from the neuromorphic dataset aspect.}
Remarkably, to cater to the event-driven paradigm which SNNs prefer, neuromorphic image datasets such as CIFAR-10-DVS \citep{Li2017CIFAR10DVSAE} and DVS-128-Gesture \citep{Amir2017ALP} have been created from dynamic vision sensors (DVS) \citep{lenero20113}.
DVS operates in an event-driven manner, only transmitting information when there is a change in the scene (pixel intensity changes), which is well-aligned with the spiking nature of SNNs.
SNNs have demonstrated outstanding performance \citep{Zhou2022SpikformerWS,yao2023spike} upon these neuromorphic datasets, showing their potential to be not only conceptual for energy efficiency but also powerful and competitive in pursuing state-of-the-art results with built-in temporal information.
However, the acquisition of dynamic image datasets for such evaluations is often encumbered by high costs and logistical inconveniences, thereby posing challenges in aligning with the pragmatic requirements of real-world applications.

% Time-series forecasting, ... \yansen{introduce the time-series forecasting task, 1. realistic data, 2. event-driven}
Acknowledging the mismatch between the preferable data format of SNNs and practical needs, we identify time-series forecasting as the potential ideal task. 
Time-series forecasting, a vital aspect of realistic data analysis including traffic \citep{li2017diffusion}, energy \citep{lai2018modeling}, etc., aims to predict future values based on historical observations arranged chronologically. 
Addressing this task often involves modeling the temporal dynamics, resonating profoundly with the nature of neural coding.

% Real-world time-series data, such as traffic data \citep{li2017diffusion} and electricity data \citep{lai2018modeling}, often exhibit trends, seasonality, and random fluctuations, requiring a thorough understanding of these components for accurate forecasting.
% However, few works have demonstrated the efficacy of SNNs in real-world multi-variant time-series forecasting.

Although SNNs are effective at managing temporal information, applying them to time-series forecasting tasks remains insufficiently explored due to some significant challenges.
% \yansen{discuss about the challenges. This should be aligned with our main contribution, e.g 1. temporal alignment between data and spiking period. 2. spike encoding / decoding for ts data, 3. unknown proper model selection}
Firstly, achieving effective temporal alignment between continuous time-series data and the discrete spiking periods of SNNs poses a hurdle, requiring careful consideration of encoding mechanisms.
% The discrete property of SNN arise from the non-trivial task of selecting an appropriate representation for temporal information in the form of spikes, necessitating robust mechanisms to mitigate information loss and noise when converting meaningful continuous values to spike trains.
A substantial disparity exists between the discrete characteristics of spike values in SNNs and the floating-point attributes of time-series data, necessitating robust mechanisms to mitigate information loss and noise when converting meaningful floating-point values to spike trains.
Moreover, the lack of standardized guidelines for proper model selection further complicates the task, calling for a thorough exploration of SNN architectures and their parameters tailored to the specific characteristics of diverse time-series datasets.
% Some studies have tried to apply SNNs in forecasting data of certain domains, such as financial data \citep{reid2014financial} and wind power data \citep{gonzalez2022spiking}, but they focus on how to deploy SNNs with simple architectures on neuromorphic chips in real-world scenarios.
% Besides, there are works addressing time-series forecasting using spike neuron P system \citep{liu2021gated, long2022multivariate}, which is not an SNN but a distributed and parallel computing paradigm.

In this paper, we propose a framework for SNNs in time-series forecasting tasks.
Firstly, by leveraging the efficiency of spiking neurons in processing time sequential information, we successfully align the time steps between time-series data and SNNs.
Secondly, we design two types of encoding layers to transfer continuous time-series data to meaningful spike trains.
Finally, we modify three types of ANNs (CNNs, RNNs, and Transformers) to their SNN counterparts with no floating-point multiplication and division, aiming to offer a guideline for proper SNN model selection for time-series forecasting tasks in deep learning age.
We conduct a comprehensive evaluation of our proposed SNN models on $4$ widely-used time-series forecasting benchmarks and the results show that SNNs achieve comparable or even better results to classic ANNs with much less energy consumption.
% Worth noting that under certain settings, the SNN models can even achieve better performance than the state-of-the-art (sota) ANNs.
Furthermore, we conduct analysis experiments to show how SNNs capture temporal dependencies within time-series data and find that SNNs can indeed model the inner dynamics of time-series data.
To sum up, our contributions can be summarized as follows:

\textbf{Framework.} We propose a unified framework for SNNs in time-series forecasting tasks, including time-series data encoding, and SNN model architecture, which offers an energy-efficient and biological-plausible alternative for time-series forecasting.

\textbf{Performance.} The presented framework enables the performance of the SNN domain to achieve comparable or even superior to existing classic ANN baselines with much less energy consumption.

\textbf{Insightful Analysis}. To the best of our knowledge, this paper stands among the first to provide a thorough analysis, encompassing both model-level investigations and temporal analysis, on how deep SNNs successfully capture features within time-series data.

\vspace{-2mm}

\section{Related Work}
\subsection{Spiking Neural Networks}\label{sec:snn}
Different from traditional ANNs, SNNs utilize discrete spike trains instead of continuous floating-point values to transmit and compute information.
According to \citet{li2023brain}, SNNs can be regarded as ANNs that incorporate bio-inspired spatiotemporal dynamics and utilize spiking activation functions (e.g., spiking neurons).
Spiking neurons, such as Izhikevich neuron \citep{Izhikevich2003SimpleMO} and Leaky Integrate-and-Fire (LIF) neuron \citep{Maas1997NetworksOS}, are usually applied to generate spike trains from floating-point values by a Heaviside step function.

Due to the non-differentiability of spike neurons, backpropagation \citep{rumelhart1986learning} can not be directly applied to train SNNs.
Nowadays, there are two mainstream approaches to address this problem.
Firstly, \textbf{ANN-to-SNN conversion} \citep{ Rueckauer2017ConversionOC, hu2018spiking} aims to convert weights of a well-trained ANN to its SNN counterpart by replacing the activation function with spiking neuron layers and adding scaling rules such as weight normalization and threshold constraints.
Another popular approach is \textbf{direct training with surrogate gradients} \citep{wu2019direct}, which introduces surrogate gradients during error backpropagation, enabling the entire procedure to be differentiable.
Backpropagation through time (BPTT) \citep{Werbos1990BackpropagationTT} is suitable for this approach, which applies the traditional backpropagation algorithm to the unrolled computational graph.
In this paper, we choose direct training with surrogate gradients as our training method for its favorable attributes, namely the avoidance of an extensive number of time steps and the elimination of adjusting training objectives based on SNN architecture.

\subsection{Time-Series Forecasting}\label{sec:tsf}
Time-series forecasting plays a crucial role in data analysis, focusing on predicting future values based on historical observations.
Early approaches, such as auto-regressive integrated moving average (ARIMA) \citep{box2015time} and Gaussian Process (GP) \citep{roberts2013gaussian}, primarily rely on statistical techniques.
With the development of deep learning, methods based on convolutional neural networks (CNN) \citep{bai2018empirical,liu2022scinet,wu2022timesnet}, recurrent neural networks (RNN) \citep{zhang2017time, siami2019performance}, Transformer \citep{wu2021autoformer,zhang2022crossformer,liu2023itransformer}, and graph neural networks (GNN) \citep{yu2017spatio,fang2023learning} have achieved great success on time-series forecasting task.
Among these approaches, GNN-based methods predominantly focus on the spatial dimension rather than the temporal aspect, diverging from the emphasis of our proposed framework for SNNs.
Consequently, our method exclusively involves the adaptation of CNNs, RNNs, and Transformers to their corresponding SNN counterparts.

Some studies have tried to apply SNNs in forecasting data of certain domains, such as financial data \citep{reid2014financial}, wind power data \citep{gonzalez2022spiking}, and electricity data \citep{kulkarni2013spiking}.
However, they either focus on how to deploy SNNs on neuromorphic hardware in real-world scenarios or fail to obtain satisfying performance due to simple architectures.
Besides, there are works addressing time-series forecasting using spike neuron P system \citep{liu2021gated, long2022multivariate}, which is not an SNN but a distributed and parallel computing paradigm.

\section{Methodology}

\subsection{Preliminaries}
\subsubsection{Task Formulation}
We consider the regular time-series forecasting task where all the time series are sequences sampled from underlying continuous signals $\mathbf{\mathcal{X}}(t)$ with constant discretization step size $\Delta T$ as $\mathbf{x}_k = \mathbf{\mathcal{X}}(k \Delta T)$.
Given the historical observed time series $\mathbf{X} = \{\mathbf{x}_1, \mathbf{x}_2, \dots, \mathbf{x}_T\}\in\mathbb{R}^{T\times C}$ for $T$ time steps, the multivariate time-series forecasting task aims to predict the values in the subsequent $L$ time steps $\mathbf{Y} = \{\mathbf{x}_{T+1}, \mathbf{x}_{T+2}, \dots, \mathbf{x}_{T+L}\}\in\mathbb{R}^{L\times C}$, where $C$ denotes the number of variates.

\subsubsection{Spiking Neurons and Surrogate Gradients}
\begin{figure}[]
    \centering
    \includegraphics[width=0.4 \textwidth]{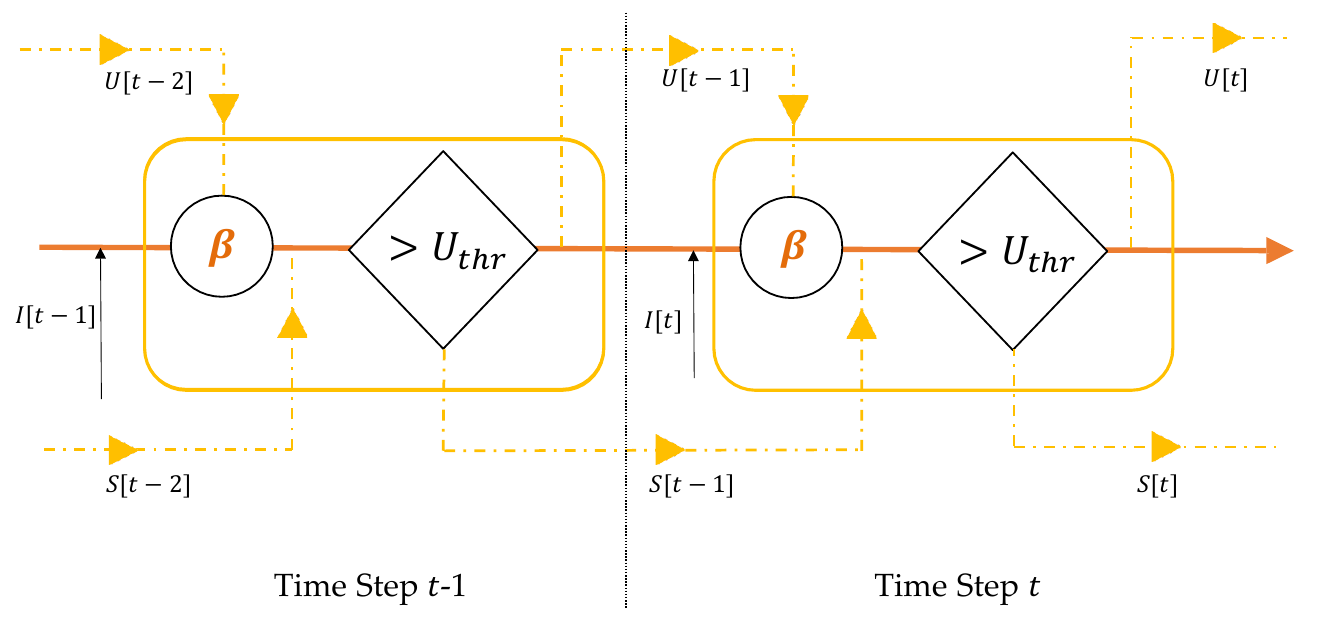}
    
    \caption{
    \label{fig:neuron}
    A recurrent representation of a leaky integrate-and-fire (LIF) neuron.
    The membrane potential $U(t-1)$ and spike $S(t-1)$ at time step $t-1$ are derived from their counterparts at time step $t-2$ and undergo processing to yield $U(t)$ and $S(t)$ at time step $t$.
    }
\end{figure}

The basic unit in SNNs is the leaky integrate-and-fire (LIF) neuron \citep{Maas1997NetworksOS} which operates on an input current $I(t)$ and contributes to the membrane potential $U(t)$ and the spike $S(t)$ at time $t$.
The dynamics of the LIF neuron shown in Figure \ref{fig:neuron} can be written as:
\begin{align}
\label{equ:membranePotential}
&U(t)=H(t-\Delta t)+I(t), \quad I(t)=f(\mathbf{x}; \mathbf{\theta}), \\
\label{equ:ht}
&H(t)=V_{reset}S(t) +\left(1-S(t)\right)\beta U(t),\\
\label{equ:heaviside}
& S(t)=
    \begin{cases}
    1, & \text{if  $U(t) \geq$ $U_{\rm thr}$} \\ 
    0, & \text{if  $U(t) <$ $U_{\rm thr}$} 
    \end{cases},
\end{align}
where $I(t)$ is the spatial input to the LIF neuron at time step $t$ calculated by applying function $f$ with $\mathbf{x}$ as input and $\mathbf{\theta}$ as learnable parameters. $H(t)$ is the temporal output of the neuron at time step $t$ and $\Delta t$ is the discretization constant controlling the granularity of LIF modeling.
The spike $S(t)$ is defined as a Heaviside step function depending on the membrane potential. When $U(t)$ achieves the threshold $U_{thr}$, the neuron will fire and emit a spike, then the temporal output $H(t)$ will be reset to $V_{reset}$. Otherwise, no spike will be emitted and the membrane potential $U(t)$ will decay to $H(t)$ by a decay rate $\beta$.

Now we generate the spike trains $\mathbf{S}\in \mathbb{R}^{T'\times N}$ with a spiking neuron layer $\mathcal{SN}(\cdot)$:
\begin{align}
    \mathbf{S} = \mathcal{SN}(\mathbf{I})
\end{align}
by iterating $T'$ steps over $N$ input currents $\mathbf{I}\in\mathbb{R}^{T'\times N}$ with $N$ LIF neurons. 

As mentioned in Section \ref{sec:snn}, we choose direct training with surrogate gradients as our method to train SNNs.
we follow \citet{SpikingJelly} to choose the arctangent-like surrogate gradients as our error estimation function when backpropagation, which regards the Heaviside step function (Equation \ref{equ:heaviside}) as:
\begin{equation} \label{equ:surrogate}
% \small
\begin{aligned}
S(t) \approx \frac{1}{\pi} \arctan(\frac{\pi}{2}\alpha U(t)) + \frac{1}{2}
\end{aligned}
\end{equation}
where $\alpha$ is a hyper-parameter to control the frequency of the arctangent function.
Therefore, the gradients of $S$ in Equation \ref{equ:surrogate} are 
$\frac{\partial S(t)}{\partial U(t)}  =\frac{\alpha}{2} \frac{1}{(1+(\frac{\pi}{2}\alpha U(t))^{2})}$
and thus the overall model can be trained in an end-to-end manner with back-propagation through time (BPTT).
% For more detailed information about training SNNs based on back-propagation through time (BPTT), please refer to Appendix \ref{sec:appendix BPTT}.

\begin{figure*}[ht]
    \centering
    \includegraphics[width=0.9\textwidth]{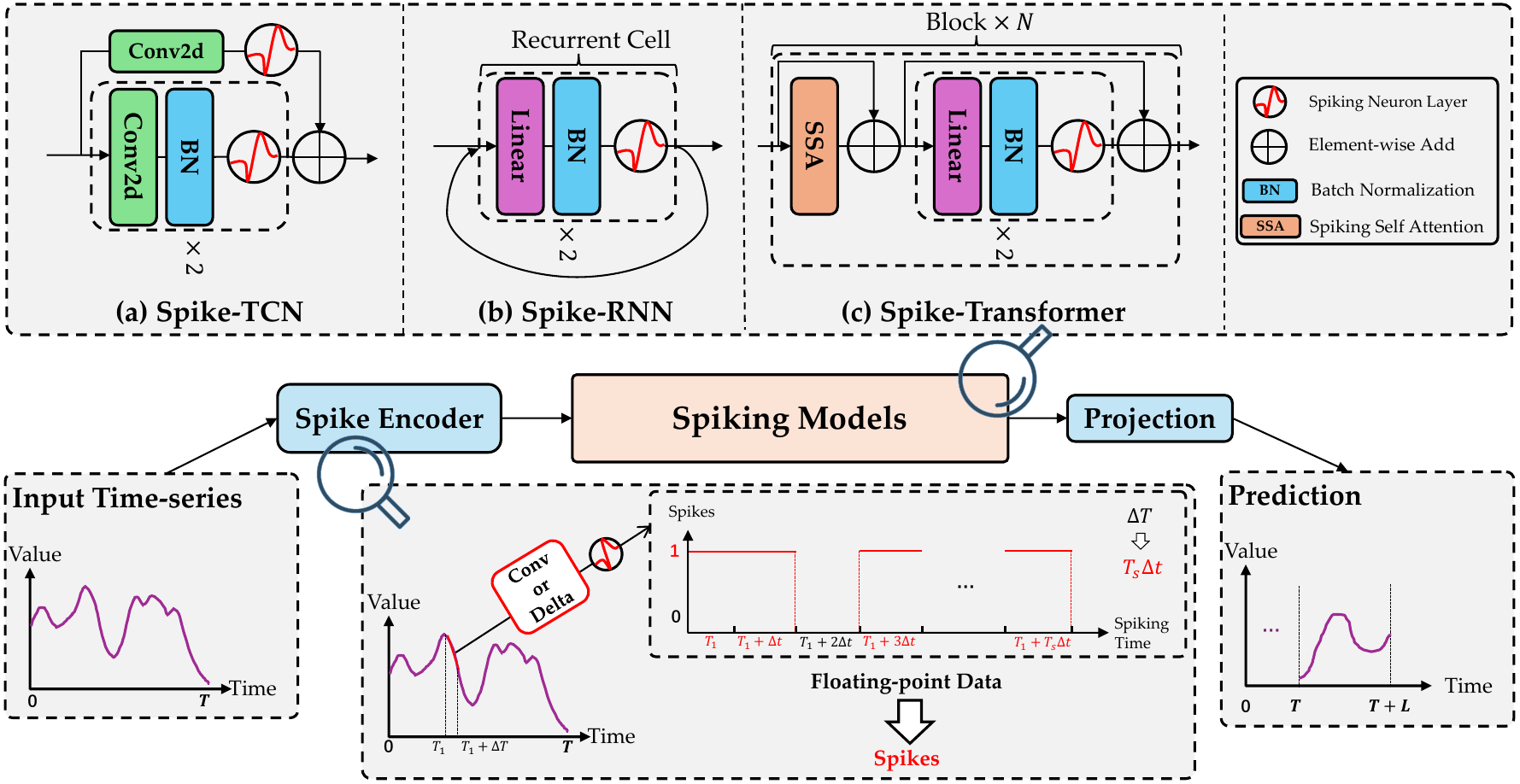}
     \vspace{-3mm}
    \caption{
    An overview of our framework for SNNs in time-series forecasting.
    Given an input time-series sample $\mathbf{X} = \{\mathbf{x}_1, \mathbf{x}_2, \dots, \mathbf{x}_T\}$ with $T$, our goal is to predict the values in the following $L$ time steps $\mathbf{Y} = \{\mathbf{x}_{T+1}, \mathbf{x}_{T+2}, \dots, \mathbf{x}_{T+L}\}$.
    Firstly, a spike encoder will be used to generate spike trains with $T_s$ spiking time steps from the original data every $\Delta t$ time step.
    After being encoded, time-series data will be converted to spike trains ($B \times T_s \times T \times C$) and will be fed into SNNs.
    We provide three SNNs: (a) Spike-TCN; (b) Spike-RNN; and (c) Spike-Transformer.
    Finally, the spike trains will be converted to floating-point values by a projection layer.
    }
    \label{fig:method}
\end{figure*}

\subsection{Temporal Alignment and Spike Encoder}\label{sec:spike encoder}
% To align the temporal dimension between time-series data and SNNs, we investigate two approaches to encode spike trains from continuous time-series information.
To utilize the intrinsic nature of SNN to its best, it's crucial to align the temporal dimension between time-series data and SNNs.
Our central concept is to incorporate relevant finer information of the spikes within the time-series data at each time step.
Specifically, we divide a time step $\Delta T$ of the time series into $T_s$ segments and each of them allows a firing event for neurons whose membrane potentials surpass the threshold, i.e., $\Delta T = T_s \Delta t$.
% \begin{align}
%     \Delta T = T_s \Delta t.
% \end{align}

This equation bridges between a time-series time step $\Delta T$ and an SNN time step $\Delta t$.
As a result, the independent variable $t$ in time-series ($\mathcal{X}(t)$) and in SNN ($U(t), I(t), H(t), S(t)$) are now sharing the same meaning.
% To this end, the spiking encoder which generates the first layer of spike trains given the float-point inputs needs to calculate $T_s \times T \times C$ possible spike events.
% The simplest non-parametric approach to treat the input time series as the current and replicate each data point $T_s$ times, however, will break the continuous hypothesis of $\mathcal{X}(t)$.
To this end, the spiking encoder, responsible for generating the first spike trains based on the floating-point inputs, needs to calculate $T_s \times T \times C$ possible spike events.
The most straightforward non-parametric approach is to consider each data point in the input time series as the current value and replicate it $T_s$ times.
However, this approach can disrupt the continuous nature of the underlying $\mathcal{X}(t)$ hypothesis.
Therefore, we seek to use parametric spike encoding techniques. 
% which also effectively serves to ``unsqueeze'' an additional dimension for the processing of spikes within SNNs.

% Rather than replicating input data for $T_{s}$ (representing time steps in SNNs) iterations, 

\paragraph{Delta Spike Encoder}
The delta spike encoder, originated from the delta modulation \cite{Eshraghian2021TrainingSN}, is inspired by the biological notion that neurons are sensitive to temporal changes. 
% Different from the original delta modulation, our delta spike encoder uses learnable threshold, which can be formulated as:
% Another avenue we explore is the use of a Delta-based Encoder.
% This approach focuses on capturing temporal differences between timestamps, emphasizing the change in information over time.
The mathematical expression governing this process is encapsulated as follows:
\begin{equation}
\mathbf{S} = \mathcal{SN}\left(\operatorname{BN}\left(\operatorname{Linear}\left(\mathbf{x}_{t}-\mathbf{x}_{t-1}\right)\right)\right)
\end{equation}
where a linear layer is applied to the temporal differences to learn different sensitivities on different SNN time steps and expand the dimension of the spike train $\mathbf{S}$ to $T_s \times T \times C$.
% $\mathbf{x}_{i+\Delta t}$ and $\mathbf{x}_{i}$ represent the data at two different timestamps $i+\Delta t$ and $i$, respectively.
% The temporal difference between these timestamps ($\Delta t$) is used as input for the linear layer.
The result undergoes batch normalization ($\operatorname{BN}$) and passed through a spiking neuron layer $\mathcal{SN}$ to be converted to spike trains.

\paragraph{Convolutional Spike Encoder}
In time-series tasks, the shapes of the sequence are often categorized as interpretable features \citep{ye2009time}  for time-series classification and clustering.
% Shapelet \citep{ye2009time} is used to find a set of sub-sequences that capture interpretable feature-based shapes.
% However, shapelet is deemed unsuitable for neuromorphic hardware implementations owing to its substantial reliance on floating-point computations.
Recently, \citet{qu2024cnn} demonstrated that this kind of morphological information could be modeled by a particular type of CNN kernel.
Therefore, we propose to use a convolutional layer as a suitable temporal encoder which should emit spikes as long as the shape of the original subsequences matches the kernel.

Given the historical observed time-series $\mathbf{X}\in\mathbb{R}^{T\times C}$, we feed it into a convolutional layer followed by batch normalization and generate the spikes as:
\begin{equation}
\mathbf{S} = \mathcal{SN}\left(\operatorname{BN}\left(\operatorname{Conv}\left(\mathbf{X}\right)\right)\right).
\end{equation}
% unsqueezing and then reshape it to meet the requirements of the following convolutional layer.
Similar to the delta spike encoder, by passing through the convolutional layer, the dimension of the spike train $\mathbf{S}$ is expanded to ${T_{s}\times T\times C}$. Spikes at every SNN time step are generated by pairing the data with different convolutional kernels.
% where $T_{s}$ represents the time steps for SNNs to processing spike trains.
% This transformation can be expressed as follows:
% where $\mathcal{SN}\left( \cdot \right)$ denotes the spiking neuron layer, and $\operatorname{BN}$ denotes batch normalization.

Both the delta spike encoder and the convolutional spike encoder capture internal temporal information of the input data, i.e., temporal changes and shapes, respectively, contributing to the representation of the dynamic nature of the information over time and catering to the following spiking layers for event-driven modeling.

% This sequence of operations facilitates the extraction of temporal patterns and changes in the input data, contributing to a comprehensive representation of the dynamic nature of the information over time.
% To sum up, in the CNN-based spike encoder, the $\Delta t$ depends on the kernel size, while in the Delta-based spike encoder, $\Delta t$ is a hyper-parameter.

\subsection{Spiking Model Architrcture} \label{sec:snn architecture}
In this section, we discuss the temporal spiking neural network to model the obtained spike trains.
We convert three distinct types of classic yet powerful temporal-oriented ANNs designed for time-series forecasting tasks, i.e., TCN, RNN (and the GRU variant), and iTransformer, to their respective SNN counterparts.

\textbf{Spike-TCN} \quad
Temporal Convolutional Network (TCN) \cite{bai2018empirical} uses convolutional kernels to model time series.
Unlike general CNNs, TCN can map any length of the time series to the same length without information leakage from the future to the past. Inspired by practice from image classification \cite{he2016deep}, recent TCN also involves the residual connection to overcome the unstable gradient problems.

Following \citet{hu2018spiking}, we construct the Spike-TCN by making the following changes to the original TCN:
1) We replace the ReLU activation function with a spiking neuron layer.
This substitution is a characteristic feature of SNNs, where the firing of neurons is modeled in a more biologically plausible way.
2) We remove the dropout operation which is hardware-unfriendly.
The dropout operation involves two steps: randomly zeroing some elements of the input tensor with a probability of $p$, and scaling the outputs by a factor of $\frac{1}{1-p}$.
The second step introduces division operations, which are not hardware-friendly.
3) We replace the residual shortcut in vanilla TCN with the spike-element-wise (SEW) residual module \citep{Fang2021DeepRL}, which implements identity mapping and overcomes the vanishing/exploding gradient problems in a spiking version.
% In summary, the Spike-TCN architecture is tailored for SNNs by incorporating elements that mimic the spiking behavior of neurons.
4) The down-sampling module is also converted to its spiking version, which follows SEW rules.

Since TCN only involves local convolution and doesn't track temporal state across time steps, the membrane potentials in Spike-TCN $U(t)$ are set to 0 at the beginning of every time-series time step. This makes it possible for parallel training.

\textbf{Spike-RNN} \quad
The vanilla recurrent neural network (RNN) uses its internal state to process the sequence of inputs and can output a sequence of the same length iteratively. 
We rewrite the recurrent cell of the original RNN to construct the Spike-RNN by substituting the activation function with the spiking neuron layer.
Unlike TCN, RNN tracks the temporal states and thus the membrane potential in Spike-RNN will persist across time steps.
We also modify the gated recurrent unit (GRU) \cite{cho2014learning}, a popular variant of RNN, which uses a gating mechanism to address the long-term dependency problem.
% As for the Spike-GRU model, the construction involves implementing a spiking gate mechanism.
% In contrast to the vanilla GRU where gating mechanisms are typically realized through standard activation functions, the spiking gate mechanism is achieved by introducing two surrogate functions after two linear layers
% , which can be formulated as: \yansen{If this is a novel mechanism, write the equation to explain the spiking gate mechanism}
% \begin{align}
%     \mathcal{SN}(...)
% \end{align}
% In addition to the dropout operation, layer normalization is not available due to its demand for floating-point multiplication operations in the inference stage.

\textbf{Spike-Transformer} \quad
The use of the Transformer architecture in the time-series forecasting task attracts a massive amount of attention \cite{wu2021autoformer, liu2023itransformer}, yet no consensus has been reached on what the best framework to apply the self-attention operation.
In this work, we build our spiking version of Transformer based on iTransformer \cite{liu2023itransformer} and name it ``iSpikformer'' considering two justifications:
1) iTransformer is the state-of-the-art time-series forecasting model on several public benchmarks and thus is strong enough to serve as our basis;
2) iTransformer treats the independent time series as tokens to capture multivariate correlations through the self-attention mechanism, which mainly focuses on spatial modeling across channels.
By constructing its spiking counterpart, we demonstrate that our design of spikes to model temporal dynamics is essentially orthogonal to spatial modeling and can be further boosted with relevant advancements.

Currently, there are various spiking Transformers designed for image classification tasks \cite{li2022spikeformer,Zhou2022SpikformerWS,zhou2023spikingformer,yao2023spike}.
Among them, Spikformer v2 \citep{zhou2024spikformer} based on Spikformer achieved current state-of-the-art performance on Imagenet-1k and CIFAR-10 benchmarks.
Therefore, we follow Spikformer to implement the spiking self-attention (SSA) mechanism and use it to replace the original self-attention layer in iTransformer to construct our spiking Transformer blocks.

Specifically, after the spike trains are obtained by the spike encoder detailed in \cref{sec:spike encoder}, a channel-wise spiking embedding layer will be applied as:
\begin{align}
    \mathbf{S}_{emb} = \mathcal{SN}(\operatorname{Linear}(\mathbf{S})),
\end{align}
where $\mathbf{S}_{emb}\in\mathbb{R}^{H\times C}$ are $C$ channel-wise embeddings of dimension $H$. These embeddings are afterward fed into the spiking Transformer blocks.

Note that the SNNs we have designed for time-series forecasting strictly adhere to hardware-friendly requirements.
Specifically, the inference process of the model avoids involving floating-point operations, such as multiply-and-accumulate (MAC) operations.
This design choice enables these models to be effectively deployed on neuromorphic chips, aligning with the hardware constraints and characteristics of such platforms.

\subsection{Spike Decoding}
% \yansen{discuss how we remap the spikes into forecasting sequences}
After passing through the final spiking neuron layer, we obtain spiking hidden states represented as $\mathbf{S}_{hidden}$.
In image classification tasks, a linear layer is commonly employed as a classification head to produce predictions.
Similarly, in the context of time-series forecasting, which is essentially a regression task, we transform the spiking data into forecasting sequences by applying a fully connected layer, denoted as $\mathbf{Y}=\operatorname{Linear}(\mathbf{S}_{hidden})$.
Since there are no additional floating-point operations applied to $\mathbf{Y}$ beyond this step, it can be accommodated within the framework of our design.

\section{Experiments}
\begin{table*}[t]
\centering
\caption{
\label{tab:main_result}
% \dongqi{showing that our SNN models are comparable with SOTA ANN models.}
Experimental results of time-series forecasting on $4$ benchmarks with different prediction lengths (horizons) $L$.
The best and the second-placed results are formatted in bold font and underlined format.
$\uparrow$ ($\downarrow$) indicates the higher (lower) the better.
All SNNs are equipped with a convolutional spike encoder in this table.
% In the \textbf{Avg.} column, numbers in {\color{blue}\textbf{blue}} font denote instances where the SNN demonstrates {\color{blue}\textbf{comparable}} performance with its ANN counterparts, whereas numbers in {\color{red}\textbf{red}} font indicate situations where the SNN {\color{red}\textbf{outperforms}} the performance of its ANN counterparts.
The numbers in the \textbf{Avg. Rank} column indicate the average ranking of the current row's models within each specific setting.
% By examining the \textbf{Avg.} and \textbf{Avg. Rank} column, we can find that SNNs exhibit performance that is \emph{comparable} to, and in some cases even \emph{superior} to, their ANN counterparts.
Numbers in the \textbf{Avg.} column with $^*$ indicate that a model significantly ($p<0.05$) outperforms its counterpart. 
All results are averaged across $3$ random seeds.
}
\resizebox{\linewidth}{!}{
\begin{tabular}{ccr:cccc:cccc:cccc:cccc:c:c}\toprule
\multirow{2}{*}{Method} & \multirow{2}{*}{Spike} & \multirow{2}{*}{Metric} & \multicolumn{4}{c:}{\bf Metr-la} & \multicolumn{4}{c:}{\bf Pems-bay} & \multicolumn{4}{c:}{\bf Solar} & \multicolumn{4}{c:}{\bf Electricity} & \multirow{2}{*}{\textbf{Avg.}} & \multirow{2}{*}{\textbf{Avg. Rank}$\downarrow$}\\
\cline{4-19}
& & & $6$ & $24$ & $48$ & $96$ & $6$ & $24$ & $48$ & $96$& $6$ & $24$ & $48$ & $96$& $6$ & $24$ & $48$ & $96$ \\
\cline{1-21}
% \multirow{4}{*}{\rotatebox{90}{Statistics}} 
\multirow{2}{*}{ARIMA} & \multirow{2}{*}{\xmark} & R$^2$$\uparrow$ &  $.687$ & $.441$ & $.282$ & $.265$ & $.741$ & $.723$ & $\underline{.692}$ & $.670$ & $.951$ & $.847$ & $.725$ & $.682$ & $.963$ & $.960$ & $.914$ & $.863$ & $.713$ & $7.3$\\
&  & RSE$\downarrow$ & $.575$ & $.742 $ & $.889$ & $.902$ & $.532 $ & $.548 $ & $\underline{.562}$ & $.612$ & $.202$ & $.365$ & $.588 $ & $.589 $ & $.522 $ & $.534 $ & $.564 $ & $.599$ & $.583$ & $7.3$\\ 
\cline{1-21}
\multirow{2}{*}{GP} & \multirow{2}{*}{\xmark} & R$^2$$\uparrow$ & $.685 $ & $.437 $ & $.265 $ & $.233 $ & $.732 $ & $.712 $ & $.689 $ & $.665 $ & $.944 $ & $.836 $ & $.711 $ & $.675 $ & $.962 $ & $.968 $ & $.912 $ & $.852$ & $.705$ & $8.4$\\ 
&  & RSE$\downarrow$ & $.572 $ & $.738 $ & $.912 $ & $.925 $ & $.544 $ & $\underline{.532}$ & $.577 $ & $.592 $ & $.225 $ & $.388 $ & $.612 $ & $.575 $ & $.603 $ & $.612 $ & $.633 $ & $.642$ & $.605$ & $7.6$\\
% \cline{1-20}
\hline \hline
% \multirow{10}{*}{\rotatebox{90}{ANN}}
% \multirow{2}{*}{Previous-TCN} & \multirow{2}{*}{\xmark} & R$^2$$\uparrow$ & $.820$ & $.601$ & $\underline{.455}$ & $\bf{.330}$ & $\underline{.881}$ & $\bf{.749}$ & $\bf{.695}$ & $\bf{.689}$ & $.958$ & $.773$ & $.645$ & $.572$ & $.975$ & $.969$ & $.968$ & $.961$ & $.753$ & $4.6$\\
% &  & RSE$\downarrow$ & $.446$ & $.665$ & $.778$ & $\bf{.851}$ & $\underline{.373}$ & $.541$ & $.583$ & $.587$ & $.210$ & $.488$ & $.617$ & $.676$ & $.282$ & $.313$ & $\bf{.319}$ & $\underline{.351}$ & $.505$ & $4.4$\\ 
% \cline{1-21}
\multirow{2}{*}{TCN} & \multirow{2}{*}{\xmark} & R$^2$$\uparrow$ & $.820$ & $.601$ & $\underline{.455}$ & $\bf{.330}$ & $\underline{.881}$ & $\bf{.749}$ & $\bf{.695}$ & $\bf{.689}$ & $.958$& $.871$ & $.737$ & $.661$ & $.975$  & $.973$ & $.968$ & $.962$ & $.770$ & $3.9$\\
&  & RSE$\downarrow$ & $.446$ & $.665$ & $.778$ & $\bf{.851}$ & $\underline{.373}$ & $.541$ & $.583$ & $.587$ & $.210$& $.359$ & $.513$ & $.583$ & $.282$  & $.287$ & $\bf{.319}$ & $\underline{.345}$ & $.483$ & $3.6$\\ 
\cline{1-21}
% \multirow{2}{*}{Previous-Spike-TCN} & \color{red}\multirow{2}{*}{\cmark} & R$^2$$\uparrow$ &  $.783$ & $.603$ & $\bf{.468}$ & $\underline{.326}$ & $.811$ & $.729$ & $.662$ & $.633$ & $.937$ & $.786$ & $.598$ & $.523$ & $.948$ & $.947$ & $.945$ & $.943$ & {$.728$} & $7.0$\\
% &  & RSE$\downarrow$ & $.491$ & $.665$ & $\underline{.769}$ & $\underline{.865}$ & $.469$ & $.541$ & $.625$ & $.635$ & $.259$& $.476$ & $.656$ & $.715$ & $.434$  & $.438$ & $.446$ & $.453$ & $.559$ & $6.7$ \\
% \cline{1-21}
\multirow{2}{*}{Spike-TCN} & \color{red}\multirow{2}{*}{\cmark} & R$^2$$\uparrow$ &  $.783$ & $.603$ & $\bf{.468}$ & $\underline{.326}$ & $.811$ & $.729$ & $.662$ & $.633$ & $.937$& $.840$ & $.708$ & $.650$ & $.970$  & $.963$ & $.958$ & $.953$ & $.750$ & $7.0$\\
&  & RSE$\downarrow$ & $.491$ & $.665$ & $\underline{.769}$ & $\underline{.865}$ & $.469$ & $.541$ & $.625$ & $.635$ & $.259$& $.401$ & $.541$ & $.596$ & $.333$  & $.342$ & $.368$ & $.389$ & $.518$ & $6.1$ \\ 
\hline \hline
\multirow{2}{*}{GRU} & \multirow{2}{*}{\xmark} & R$^2$$\uparrow$ & $.759$ & $.429$ & $.301$ & $.194$ & $.747$ & $.703$ & $.691$ & $.665$ & $.950$ & $.875$ & $.781$ & $\underline{.737}$ & $\bf{.981}$ & $.972$ & $.971$ & $\bf{.964}$ & $.733$ & $5.8$\\
&  & RSE$\downarrow$ & $.517 $ & $.797 $ & $.882 $ & $.947  $ & $.529 $ & $.573 $ & $.584 $ & $.608 $ & $.219 $ & $.355 $ & $.476 $ & $\underline{.522}$ & $.506 $ & $.598 $ & $.537 $ & $.587$ & $.573$ & $7.1$\\ 
\cline{1-21}
\multirow{2}{*}{Spike-GRU} & \color{red}\multirow{2}{*}{\cmark} & R$^2$$\uparrow$ & $\bf{.846}$ & $.615$ & $.427$ & $.275$ & $.864$ & $.741$ & $.688$ & $.657$ & $.912$& $.822$ & $.771$ & $.668$ & $.978$  & $.964$ & $.962$ & $.959$ & $.759$ & $6.2$\\
&  & RSE$\downarrow$ & $\underline{.414}$ & $\underline{.663}$ & $.827$ & $.943$ & $.398$ & $.535$ & $.601$ & $.621$ & $.299$ & $.430$ & $.485$ & $.629$ & $.280$ & $.317$ & $\underline{.338}$ & $.484$ & $.517^*$ & $6.0$\\ 
\cline{1-21}
\multirow{2}{*}{Spike-RNN} & \color{red}\multirow{2}{*}{\cmark} & R$^2$$\uparrow$ & $\bf{.846}$ & $\underline{.622}$ & $.433$ & $.283$ & $.872$ & $\underline{.745}$ & $.685$ & $.654$ & $.923$ & $.820$ & $\bf{.812}$ & $.714$ & $.977$ & $.972$ & $.962$ & $.960$ & $.768^*$ & $5.2$\\
&  & RSE$\downarrow$ & $\bf{.412}$ & $\bf{.648}$ & $.794$ & $.935$ & $.387$ & $\bf{.528}$ & $.588$ & $.634$ & $.278$& $.425$ & $\bf{.435}$ & $.586$ & $.267$  & $.296$ & $.346$ & $.481$ & $.503^*$ & $4.8$\\ 
% \cline{1-20}
\hline \hline
\multirow{2}{*}{Autoformer} & \multirow{2}{*}{\xmark} & R$^2$$\uparrow$ & $.762 $ & $.548 $ & $.411 $ & $.282$ & $.782$ & $.711 $ & $.689 $ & $.668 $ & $.960 $ & $.852 $ & $.791 $ & $.701 $ & $\underline{.980}$ & $\bf{.977}$ & $\bf{.975}$ & $\underline{.963}$ & $.753$ & $4.6$\\
&  & RSE$\downarrow$ & $.565 $ & $.692 $ & $.785 $ & $.872 $ & $.452 $ & $.543 $ & $.577 $ & $\bf{.565}$ & $.212 $ & $.432 $ & $.622 $ & $.685 $ & $.481 $ & $.506 $ & $.566 $ & $.548$ & $.569$ & $6.6$\\ 
\cline{1-21}
\multirow{2}{*}{iTransformer} & \multirow{2}{*}{\xmark} & R$^2$$\uparrow$ & $\underline{.829}$ & $\bf{.623}$ & $.439 $ & $.285$ & $\bf{.887}$ & $.719 $ & $.685 $ & $.668$ & $\bf{.964}$ & $\bf{.879}$ & $\underline{.799}$ & $\bf{.738}$ & $.979$ & $\bf{.977}$ & $\bf{.975}$ & $\bf{.964}$ & $\bf.776$ & $\bf{2.9}$\\
&  & RSE$\downarrow$ &$.436 $ &$\bf{.648}$ &$.780$ & $.878$ & $\bf{.362}$ & $.547$ & $\bf{.561}$ & $.584$ & $\bf{.191}$ & $\underline{.348}$ & $\underline{.448}$ & $.563$ & $\bf{.259}$ & $\underline{.305}$ & $\underline{.335}$ & $.427$ & $\underline{.480}$ & $\bf{2.8}$\\ 
\cline{1-21}
% \multirow{8}{*}{\rotatebox{90}{SNN (Ours)}} 
\multirow{2}{*}{iSpikformer} & \color{red}\multirow{2}{*}{\cmark}& R$^2$$\uparrow$ & $.817$ & $.618$ & $.440$ & $.279$ & $.879$ & $.744$ & $.687$ & $\underline{.674}$ & $\underline{.961}$ & $\underline{.876}$ & $.795$ & $\bf{.738}$ & $.977$ & $\underline{.974}$ & $\underline{.972}$ & $\underline{.963}$ & $\underline{.775}$ & $\underline{3.5}$\\
&  & RSE$\downarrow$ & $.475$ & $.668$ & $\bf{.752}$ & $.905$ & $.376$ & $.536$ & $.569$ & $\underline{.580}$ & $\underline{.204}$ & $\bf{.333}$ & $.465$ & $\bf{.521}$ & $\underline{.263}$ & $\bf{.284}$ & $.338$ & $\bf{.348}$ & $\bf.476$ & $\underline{2.9}$\\ 
% \cline{1-20}
\bottomrule
\end{tabular}
}
\end{table*}

In this section, we conduct experiments to investigate the following research questions: 

\textbf{RQ1}: Encompassing the merits of energy efficiency, biological plausibility, and the event-driven paradigm, can these SNNs achieve comparable performance to their ANN counterparts?

\textbf{RQ2}: Is our design of temporal alignment and corresponding spike encoders effective and robust?

\textbf{RQ3}: To what extent do the SNNs help reduce energy consumption?

\subsection{Experiment Settings}
To assess the forecasting capabilities of the compared methods, the datasets listed below are employed:
\textbf{Metr-la} \citep{li2017diffusion}: Average traffic speed measured on the highways of Los Angeles County;
\textbf{Pems-bay} \citep{li2017diffusion}: Average traffic speed in the Bay Area;
\textbf{Electricity} \citep{lai2018modeling}: Hourly electricity consumption measured in kWh;
\textbf{Solar} \citep{lai2018modeling}: Records of solar power production.

On these forecasting datasets, we compare our method with 
two statistics methods \textbf{ARIMA} \citep{box2015time} and \textbf{GP} \citep{roberts2013gaussian};
one CNN-based model \textbf{TCN} \cite{bai2018empirical};
one RNN-based model \textbf{GRU} \citep{cho2014learning};
two Transformer-based models \textbf{Autoformer} \citep{wu2021autoformer} and \textbf{iTransformer} \cite{liu2023itransformer}.
% It's worth noting that we have chosen to use GRU as our RNN baseline due to numerous studies \citep{torres2021deep} demonstrating that the performance of RNNs and GRUs is nearly indistinguishable in the context of time series forecasting.

For evaluating time-series forecasting tasks, we employ the Root Relative Squared Error (RSE) and the coefficient of determination (R$^2$). The detailed statistics of datasets, hyper-parameters of models, and formulation of the evaluation metrics can be referred to Appendix \ref{sec:exp settings}.

\subsection{Main Results}

We report the results of all the methods on $4$ time-series forecasting tasks with various prediction lengths $L$ in \Cref{tab:main_result}.
Results for ARIMA, GP, GRU, and Autoformer are obtained from the study conducted by \citet{fang2023learning}.

\begin{figure}[htp]
\centering
\includegraphics[width=0.98\linewidth]{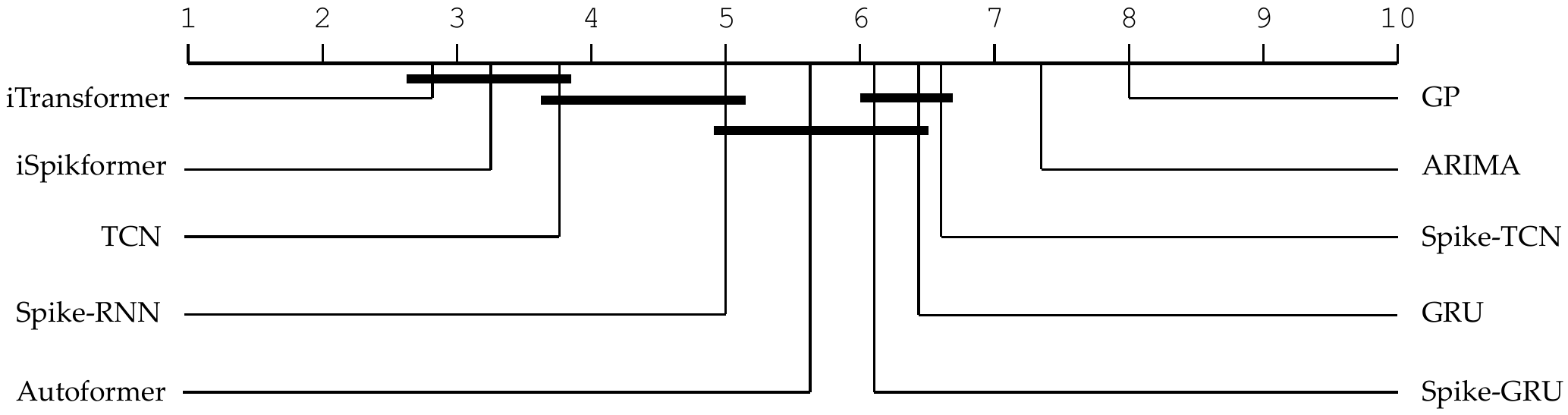}
\caption{
Critical Difference (CD) diagram of all methods in Table \ref{tab:main_result} on time series forecasting tasks with a confidence level of $95\%$.
}
\label{fig:avg rank}
\end{figure}

% Before delving into the discussion of the experimental results, we want to clarify our objective, which is to enable SNNs to attain results comparable to those of ANNs.
% This aim is rooted in the advantages of SNNs, which encompass energy efficiency, biological plausibility, and an event-driven paradigm.
% If SNNs can achieve performance on par with ANNs, they become a promising candidate for application in neuromorphic chips.
To conclude from \Cref{tab:main_result}, all the SNNs successfully model the time series and achieved reasonable performances, and our iSpikformer achieves comparable or even better performance compared to the state-of-the-art ANN model, answering \textbf{RQ1}.
To elaborate:

\textbf{(1) SNNs succeed when temporal dynamics are properly preserved and handled.}  % achieve comparable or superior performance compared to their ANN counterparts.
While Spike-TCN underperforms TCN, Spike-GRU, and Spike-RNN achieved significantly better performances over the GRU baseline. We attribute this phenomenon to the different ability of the models to handle sequential dynamics. Similar to TCN, Spike-TCN models only local features of the sequence and doesn't track the temporal states. As a result, the membrane potentials are hardly reset to 0 across time steps, which violates the nature of SNNs. On the other hand, Spike-RNN and Spike-GRU persist in such temporal information and cater to our overall event-driven paradigm. It is worth highlighting that Spike-RNN can achieve state-of-the-art performance on benchmarks with both long and short historical observations.
Furthermore, by examining the critical difference diagram depicted in Figure \ref{fig:avg rank}, we can roughly categorize all the methods we have employed into three tiers, and it was a pleasant surprise to observe that our iSpikformer ranks within the top tier.

\begin{figure*}[htp]
  \centering
    \subfigure[]{
       \centering
       \includegraphics[width=0.23 \textwidth]{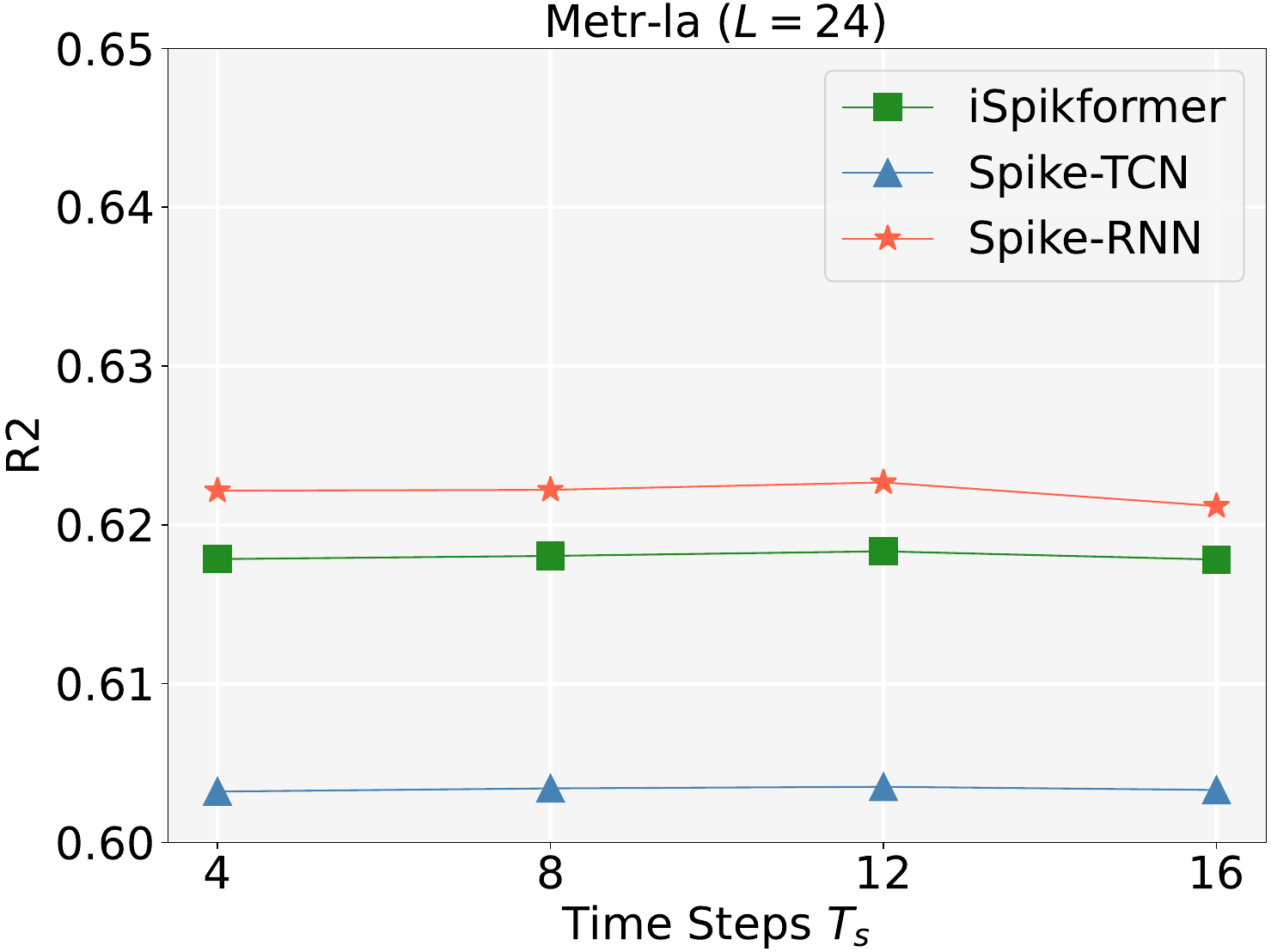}
    }
    \subfigure[]{
        \centering
        \includegraphics[width=0.23 \textwidth]{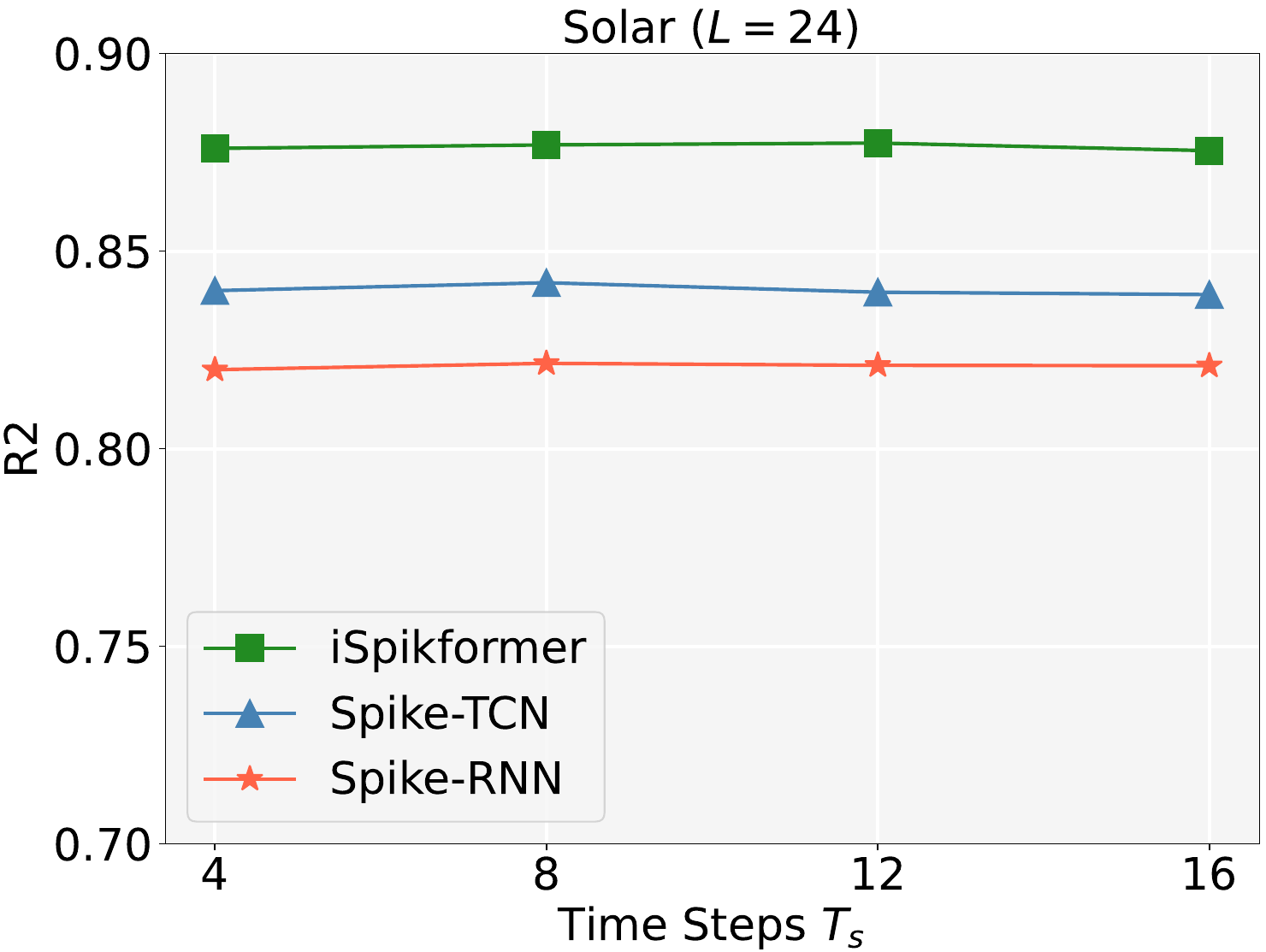}
    }
    \subfigure[]{
        \centering
        \includegraphics[width=0.23 \textwidth]{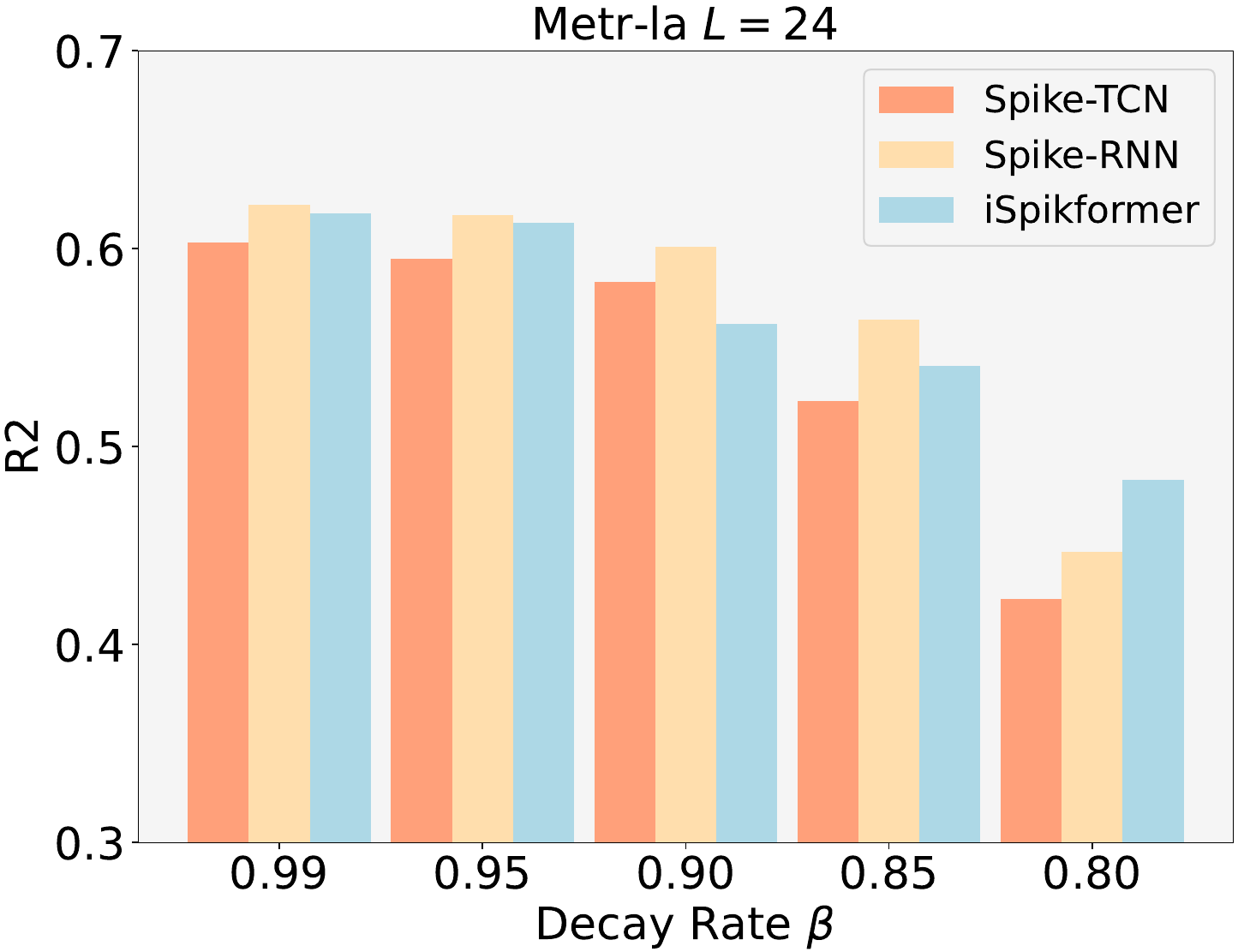}
    }
    \subfigure[]{
        \centering
        \includegraphics[width=0.235 \textwidth]{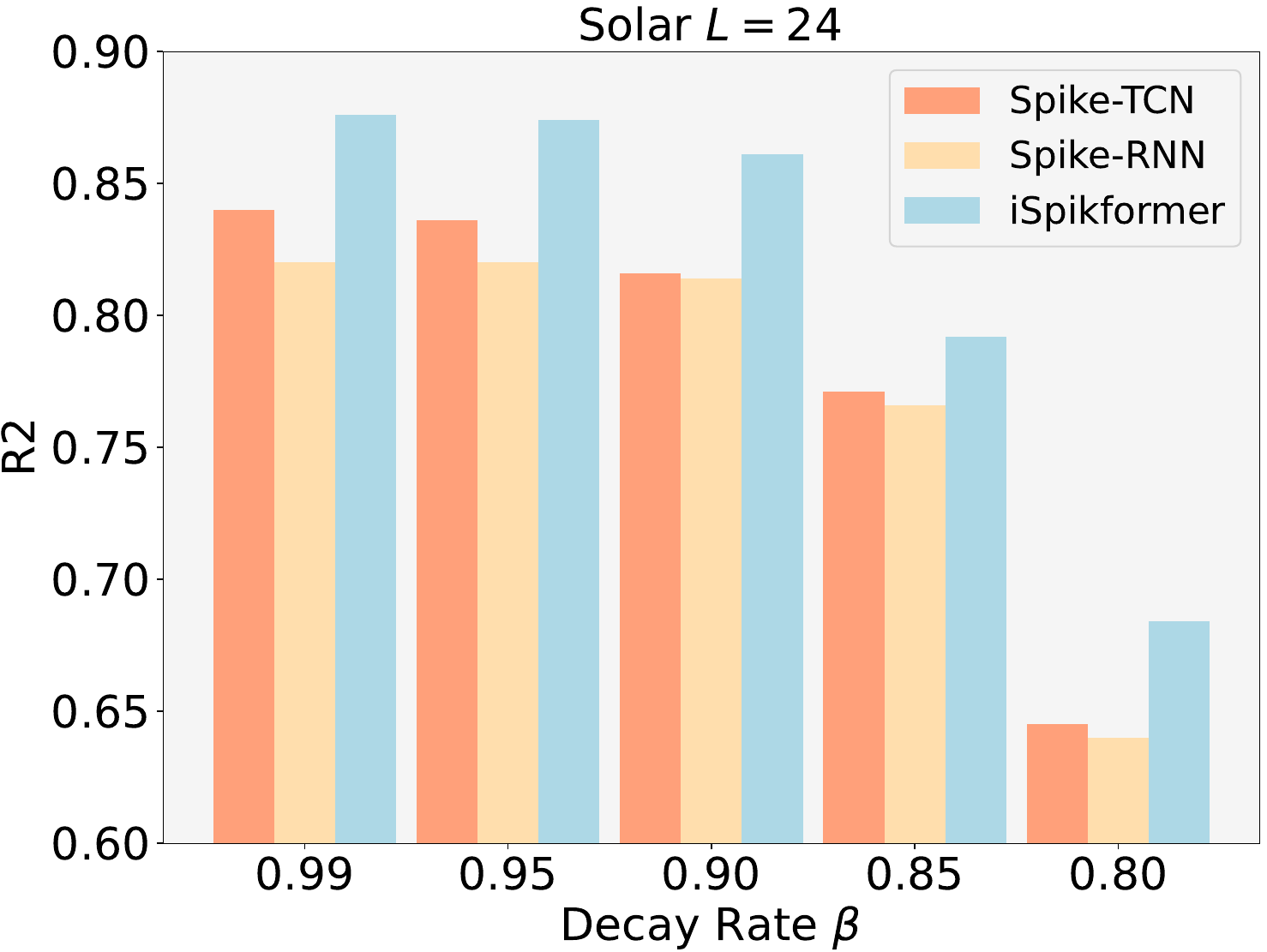}
    }
 % \vspace{-3mm}
  \caption{\label{fig:ablationstudy}
  The impact of two crucial hyper-parameters in SNNs: time Steps $T_s$ and the decay rate $\beta$.
  (a) and (b): R$^2$ versus $T_s$ on Metr-la and Solar respectively.
  (c) and (d): R$^2$ versus $\beta$ on Metr-la and Solar respectively.
  The horizon $L$ of these experiments is set to $24$.
  }
\end{figure*}

\textbf{(2) SNNs with our temporal modeling can be further improved by advanced spatial modeling techniques.}
One of the standout findings in our study is the exceptional performance of iSpikformer, which exhibits results nearly indistinguishable from the state-of-the-art ANN model, iTransformer.
As observed in Table \ref{tab:main_result}, iSpikformer achieves the lowest average RSE compared to all other methods and it nearly matches the R$^2$ performance of iTransformer, with only a marginal decrease of $0.001$.
Furthermore, it's noteworthy that the RSE of iSpikformer consistently ranks as either the lowest or the second lowest across all benchmark datasets.
These findings confirm that our temporal modeling techniques perfectly complement the most advanced spatial modeling technique, and together the spiking neural network can be very effective to address the general time-series forecasting task.

% \textbf{(5) All SNNs outperform traditional statistics methods.}
% Some studies \cite{torres2021deep,lim2021time} have proved that traditional statistics methods like ARIMA and GP tend to exhibit inferior performance compared to neural network-based methods, particularly as the forecasting horizon extends.
% Our conducted experiments also confirm this observation, revealing the potential practical applicability of our proposed SNNs across various domains of study.

\subsection{Model Analysis}
We conduct model analysis experiments to answer \textbf{RQ2}, i.e., to verify whether our proposed methods are effective and robust. %: (1) spike encoder types, (2) two temporal-aware hyper-parameters: time step $T_s$ in SNNs, and decay rate $\beta$ in Equation \ref{equ:ht}.

\subsubsection{Encoder Type}
As discussed in Section \ref{sec:spike encoder}, spike encoders are used to convert continuous time-series data to spike trains. 
To verify the effectiveness of our proposed spike encoder, we evaluate the performance of Spike-TCN, Spike-RNN, and iSpikformer with three different types of encoders: CNN-based encoder, Delta-based encoder, and Repetition encoder.
Repetition encoder refers to the most widely-used encoding methods in previous SNN studies, where the time-series data is repeated $T_s$ times to add an extra dimension.
In this experiment, we focus on Metr-la and Electricity datasets, taking into account the different lengths of historical observations.
The results of this comparison are presented in Table \ref{tab:encoder}.

\begin{table}[htp]
% \vspace{-3mm}
\centering
\small
\caption{
\label{tab:encoder}
Performance of Spike-TCN, Spike-RNN, and iSpikformer on two benchmarks with different encoder types.
Numbers with $*$ indicate that SNNs fail to converge in the settings.
Numbers presented in \textbf{bold} font indicate that this particular type of encoder has achieved the best performance within the SNN model.
}
\resizebox{0.98\linewidth}{!}{
\begin{tabular}{ccr:cccc:cccc}
\toprule
\multirow{2}{*}{Method} & \multirow{2}{*}{Encoder} & \multirow{2}{*}{Metric} & \multicolumn{4}{c:}{\bf Metr-la} & \multicolumn{4}{c}{\bf Electricity}\\
\cline{4-11}
& & & $6$ & $24$ & $48$ & $96$ & $6$ & $24$ & $48$ & $96$\\
\cline{1-11}
\multirow{6}{*}{\rotatebox{90}{Spike-TCN}} & \multirow{2}{*}{Convolutional} & R$^2$$\uparrow$ &  $\bf.783$ & $\bf.603$ & $\bf.468$ & $\bf.326$ & $\bf.970$ & $\bf.963$ & $\bf.958$ & $\bf.953$\\
&& RSE$\downarrow$ & $\bf.491$ & $\bf.664$ & $.769$ & $.935$ & $\bf.333$ & $\bf.342$ & $\bf.368$ & $\bf.389$ \\
\cline{2-11}
& \multirow{2}{*}{Delta} & R$^2$$\uparrow$ & $.751$ & $.582$ & $.458$ & $.317$ & $.963$ & $.956$ & $.948$ & $.942$\\
&& RSE$\downarrow$ & $.525$ & $.676$ & $\bf.768$ & $\bf.871$ & $.344$ & $.371$ & $.432$ & $.460$ \\
\cline{2-11}
& \multirow{2}{*}{Repetition} & R$^2$$\uparrow$ &  $.024*$ & $.024*$ & $.022*$ & $.020*$ & $.878$ & $.710*$ & $.710*$ & $.710*$ \\
&& RSE$\downarrow$ & $1.05*$ & $1.04*$ & $1.04*$ & $1.05*$ & $.662$ & $1.03*$ & $1.03*$ & $1.03*$\\

\hline \hline

\multirow{6}{*}{\rotatebox{90}{Spike-RNN}} & \multirow{2}{*}{Convolutional} & R$^2$$\uparrow$ & $\bf.846$ & $\bf.622$ & $\bf.433$ & $\bf.283$ & $\bf.977$ & $\bf.972$ & $\bf.962$ & $\bf.960$\\
&& RSE$\downarrow$ & $\bf.412$ & $\bf.648$ & $\bf.794$ & $\bf.935$ & $\bf.267$ & $\bf.296$ & $.346$ & $\bf.481$ \\
\cline{2-11}
& \multirow{2}{*}{Delta} & R$^2$$\uparrow$ & $.839$ & $.616$ & $.430$ & $.277$ & $.969$ & $.966$ & $.962$ & $.876$\\
&& RSE$\downarrow$ & $.420$ & $.652$ & $.799$ & $.938$ & $.301$ & $.318$ & $\bf.344$ & $.685$ \\
\cline{2-11}
& \multirow{2}{*}{Repetition} & R$^2$$\uparrow$ & $.817$ & $.578$ & $.021*$ & $.021*$ & $.901$ & $.816$ & $.710*$ & $.710*$\\
&& RSE$\downarrow$ & $.481$ & $.684$ & $1.04*$ & $1.04*$ & $.592$ & $.766$ & $1.03*$ & $1.04*$ \\

\hline \hline

\multirow{6}{*}{\rotatebox{90}{iSpikformer}} & \multirow{2}{*}{Convolutional} & R$^2$$\uparrow$ & $\bf.817$ & $\bf.618$ & $\bf.440$ & $\bf.279$ & $\bf.977$ & $\bf.974$ & $\bf.972$ & $\bf.963$\\
&& RSE$\downarrow$ & $\bf.475$ & $.668$ & $\bf.752$ & $\bf.905$ & $\bf.263$ & $\bf.284$ & $\bf.338$ & $\bf.348$ \\
\cline{2-11}
& \multirow{2}{*}{Delta} & R$^2$$\uparrow$ & $.804$ & $.601$ & $.434$ & $.272$ & $.972$ & $.969$ & $.960$ & $.944$\\
&& RSE$\downarrow$ & $.496$ & $\bf.666$ & $.759$ & $.910$ & $.274$ & $.302$ & $.391$ & $.455$ \\
\cline{2-11}
& \multirow{2}{*}{Repetition} & R$^2$$\uparrow$ & $.692$ & $.548$ & $.238$ & $.021*$ & $.962$ & $.953$ & $.849$ & $.710*$\\
&& RSE$\downarrow$ & $.573$ & $.708$ & $.847$ & $1.04*$ & $.289$ & $.557$ & $.705$ & $1.03*$\\
\bottomrule
\end{tabular}
}
\end{table}

In summary, our proposed spike encoders, including the convolutional spike encoder, and delta spike encoder, have demonstrated their effectiveness in capturing temporal information from time-series data.
Specifically:
(1) While repetition is a common strategy utilized in many tasks, it appears that SNNs with repetition encoders may struggle to converge under many settings in time-series forecasting tasks. This confirms the necessity of an exquisite temporal modeling technique.
(2) Both convolutional spike encoder and delta spike encoder are event-driven spike generators, however, the performance of SNNs utilizing convolutional spike encoders surpasses that of SNNs employing delta spike encoders by an increase of $0.09$ on average. This improvement demonstrates that the shape-based encoder which takes a wide scope of sequence into consideration is more effective than a change-based encoder which reacts to only the very local changes.

\subsubsection{Hyper-parameters}
To verify the robustness of our design, we investigate how sensitive the performances respond to different choices of time-related hyper-parameters.

\textbf{Time Step.} \quad
As introduced in Section \ref{sec:spike encoder}, time step $T_s$ in SNNs is a hyper-parameter that controls how accurately the SNNs model the temporal dynamics $\Delta t$.
% In theory, as $T_s$ increases, the performance should improve due to the more detailed modeling of temporal dynamics.
We conduct experiments with varying values of $T_s$ from the set $\left\{4, 8, 12, 16\right\}$ on the Metr-la and Solar benchmarks, both with a horizon of $L=24$.
Based on the results presented in Figure \ref{fig:ablationstudy} (a) and (b), we observe that, in general, the R$^2$ values remain relatively stable with minimal variation as $T_s$ increases.
However, there may be a slight decrease in R$^2$ when $T_s = 16$.
This phenomenon can be reasonably explained by the occurrence of ``\emph{self-accumulating dynamics}'' \citep{Fang2020IncorporatingLM}.
This refers to an error accumulation triggered by surrogate gradients, which can lead to gradient vanishing or explosion, potentially affecting the model's performance adversely.

\textbf{Decay Rate.} \quad
% $\beta$ in Equation \ref{equ:ht} is configured to regulate the rate of decay for the membrane potential $U[t]$ at time step $t$ in cases where no spikes occur at time step $t-1$.
We performed experiments using different values of $\beta$ taken from the set $\left\{0.99, 0.95, 0.90, 0.85, 0.80\right\}$ on the Metr-la and Solar benchmarks, both with a forecasting horizon of $L=24$.
As illustrated in Figure \ref{fig:ablationstudy} (c) and (d), it is evident that when the value of $\beta$ increases, there is a noticeable decrease in R$^2$ performance on both the Metr-la and Solar benchmarks.
This observation can be attributed to the fact that a higher $\beta$ makes the SNN more persistent in its internal state, which is beneficial for retaining long-term information.

\subsection{Energy Reduction}

An essential advantage of SNNs is the low consumption of energy during inference. 
Assuming that we run Spike-TCN, Spike-RNN, and iSpikformer on a 45nm neuromorphic hardware \cite{horowitz20141}, we can calculate the theoretical energy consumption (Appendix \ref{sec:appendix energy consumption}).
We compare the theoretical energy consumption per sample of our proposed three SNNs and their original ANN counterparts on test sets of Electricity benchmarks with $L=24$ during inference (Table \ref{tab:energy}).

\begin{table}[htp]
% \vspace{-3mm}
\small
\begin{center}
\caption{\label{tab:energy}
Theoretic energy consumption per sample of Electricity during the inference stage.
``OPs'' refers to SOPs in SNN and FLOPs in ANN.
``SOPs'' denotes the synaptic operations of SNNs.
``FLOPs'' denotes the floating point operations of ANNs.
}
\vspace{-3mm}
\resizebox{ \linewidth}{!}{
\begin{tabular}{c|c|c|c|c|c}
\toprule
\hline
\bf Model & \bf Param(M) &\bf OPs ($\bf G$) &\bf Energy ($\bf mJ$) & \bf Energy Reduction &\bf R$^2$ \\
\hline

TCN & $0.460$ & $0.14$ & $0.64$ & {\multirow{2}{*}{$\bf 63.60\% \downarrow$ }} & $.973$  \\
Spike-TCN & $0.461$ & $0.15$ & $0.23$ &  & $.963$ \\ 
\hline

GRU & $1.288$ & $1.32$ & $6.07$ & {\multirow{2}{*}{$\bf 75.05\% \downarrow$ }} & $.972$  \\
Spike-GRU & $1.289$ & $1.63$ & $1.51$ &  & $.964$ \\
\hline

iTransformer & $1.634$ & $2.05$ & $9.47$ & {\multirow{2}{*}{$\bf 66.30\% \downarrow$ }} & $.977$  \\
iSpikformer & $1.634$ & $3.55$ & $3.19$ &  & $.974$ \\ 
\bottomrule
\hline

\end{tabular}
}
\end{center}
\end{table}

As shown in Table \ref{tab:energy}, SNNs exhibit a notably lower energy consumption compared to their ANN counterparts, resulting in an average decrease of approximately $70.33\%$, which answers \textbf{RQ3}.
This compelling advantage positions SNNs as an attractive choice for energy-efficient solutions.
It is worth emphasizing that the parameters of an SNN and an ANN are nearly identical, as we only eliminate certain parameter-free operations, such as layer normalization and dropout when converting an ANN to its SNN version.
Notably, Spike-GRU demonstrates a significant reduction in energy consumption, reaching up to $75.05\%$ less compared to the traditional GRU model. 
This observation suggests that Spike-GRU exhibits the lowest average firing rate of spiking neurons and the most sparse data flow among the considered models, contributing to its energy efficiency.

\subsection{Temporal Analysis}
\label{sec:temporal analysis}

\begin{figure}[htp]
  \centering
    \subfigure[]{
       \centering
       \includegraphics[width=0.19 \textwidth]{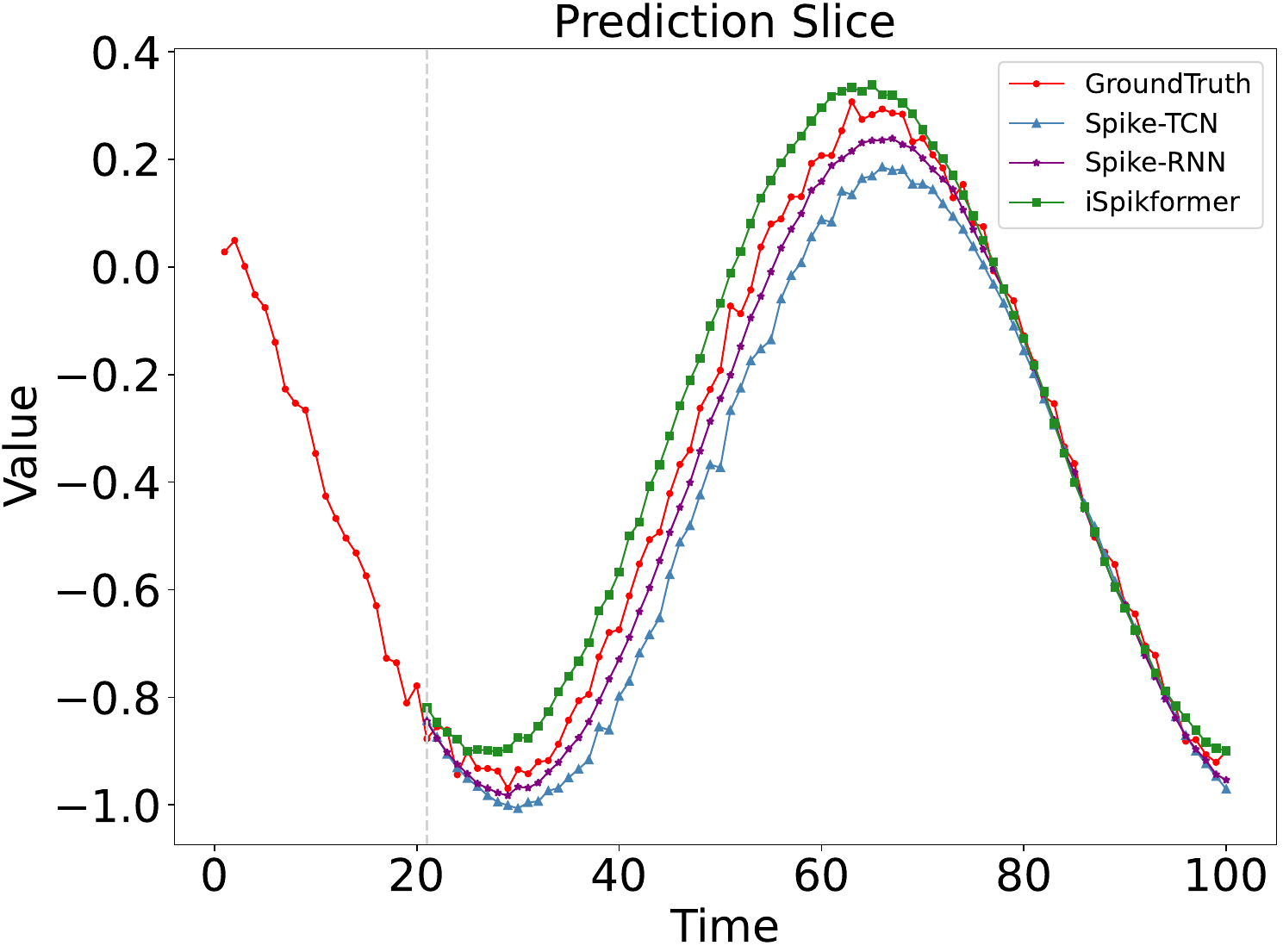}
    }
    \subfigure[]{
        \centering
        \includegraphics[width=0.23 \textwidth]{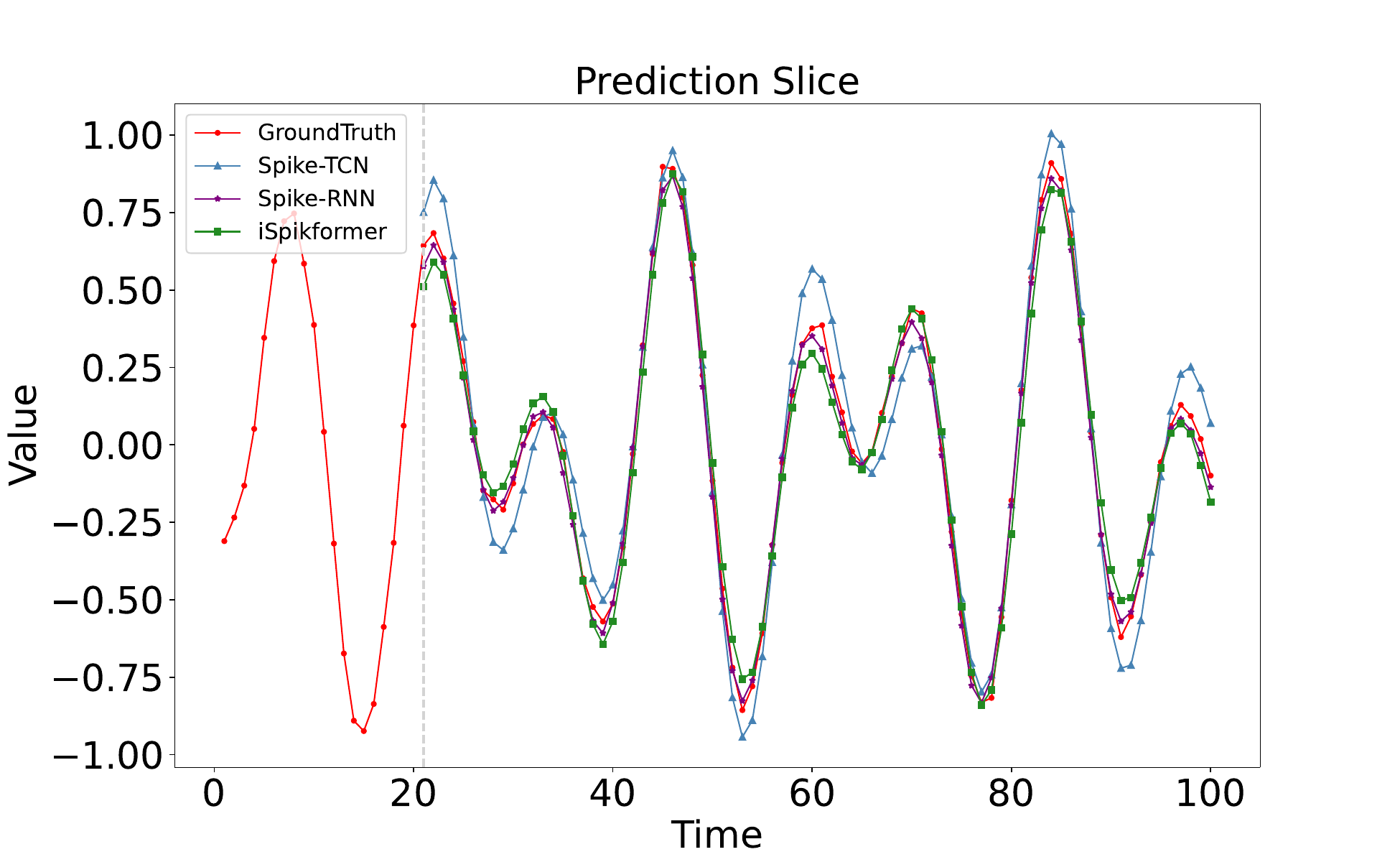}
    }
     % \vspace{-3mm}
  \caption{\label{fig:temporal analysis}
  A prediction slice ($T=20, L=80$) of Spike-TCN, Spike-RNN, and iSpikformer on synthetic time-series data.
  (a) Prediction slice on low-frequency data.
  (a) Prediction slice on high-frequency data.
  }
\end{figure}

To validate the capability of the proposed SNNs for time series prediction in capturing the time variation characteristics of time series data, we conducted experiments on both low-frequency and high-frequency \emph{synthetic periodic signals} using Spike-TCN, Spike-RNN, and iSpikformer, as illustrated in Figure \ref{fig:temporal analysis}. Details of the synthetic periodic signals can be found in \Cref{sec:appendix synthetic}

Based on the observations from Figure \ref{fig:temporal analysis}, we can draw the following conclusions:
(1) Spike-TCN, Spike-RNN, and iSpikformer all demonstrate strong performance when dealing with both low-frequency and high-frequency time-series data;
(2) When dealing with high-frequency signals, Spike-TCN seems to exhibit lower accuracy in predicting peak values compared to Spike-RNN and iSpikformer.
This observation could potentially account for the relatively weaker overall performance of Spike-TCN within our proposed SNNs.
% (1) Spike-TCN, Spike-RNN, and iSpikformer all demonstrate strong performance when dealing with low-frequency time-series data;
% (2) In the case of high-frequency time series, we notice that Spike-RNN and iSpikformer tend to excel at predicting the trends within the time series. On the other hand, Spike-TCN appears to focus more on modeling the dynamic aspects of the data.
% This finding aligns with the conclusions put forth by \citet{yue2022ts2vec}, indicating the efficiency of our proposed SNNs in time-series forecasting.
Refer to \Cref{app:case study} for visualizations of forecasting results on real datasets. 

\section{Conclusion}
% \textbf{Limitations.}
% The limitations of our work can be summarized as follows.
% In the case of SNNs for classification tasks, the use of a fully connected layer for predicting classification results is common. The classification results can be derived by counting the firing rate of spike trains within each classification category.
% Consequently, previous research often omitted the spiking neuron layers added to the projection layer.
% However, in regression tasks like time-series forecasting, we rely on the floating-point output of the final projection layer.
% In future work, we aim to explore ways to bridge this gap and potentially find methods that allow for a more unified approach in both classification and regression tasks within the SNN framework.

% \textbf{Future directions.}
% Future research directions encompass:
% (1) Developing hardware-friendly algorithms that enable parallel training of TCN-like models without the need to reset $U(t)$ to $0$.
% (2) Exploring more promising methods to enhance the utilization of SNNs for capturing spatial information within time-series data.

% \textbf{Conclusion.}
In this paper, we present a framework for utilizing SNNs in time-series forecasting tasks.
Through a series of experiments, we have demonstrated the efficacy of our proposed SNN-based approaches in time-series forecasting.
These approaches have shown comparable performance to traditional time-series forecasting methods across diverse benchmark datasets, while significantly reducing energy consumption.
Furthermore, our detailed analysis experiments have shed light on the SNN's ability to capture temporal dependencies within time-series data.
This insight underscores the nuanced strengths and effectiveness of SNNs in modeling the intricate dynamics of time series.
In summary, our study contributes to the expanding field of SNNs, offering an energy-efficient and biologically plausible alternative for time-series forecasting tasks.
The limitations and future directions are discussed in Appendix \ref{app:limit}.

\section*{Impact Statement}

Our research contributes to the growing field of spiking neural networks and presents a promising alternative for time-series forecasting tasks.
It paves the way for the development of more biologically inspired and temporally aware forecasting models, offering exciting prospects for future advancements in this domain.
Besides, we do not think our work will have a bad impact on ethical aspects and future societal consequences.

\section*{Acknowledgments}
The authors would like to thank the anonymous reviewers for their valuable comments.
This work was supported partially by National Natural Science Foundation of China (No. 62076068).

\bibliography{main}
\bibliographystyle{icml2024}

\clearpage
\appendix
\section{Experiment Settings}\label{sec:exp settings}
\subsection{Statistics of Datasets}
We partitioned the forecasting datasets into train, validation, and test sets following a chronological order.
The statistical characteristics and specific split details can be found in Table \ref{tab:exp setting}.

\begin{table}[h]
\centering
\resizebox{\linewidth}{!}{
\begin{tabular}{lcccc}
\hline
Dataset & Samples & Variables & Length & Train-Valid-Test Ratio \\
\hline
Metr-la & $34,272$ & $207$ & $12$ & $(0.7, 0.2, 0.1)$ \\
Pems-bay & $52,116$ & $325$ & $12$  & $(0.7, 0.2, 0.1)$ \\
Solar-energy & $52,560$ & $137$ & $168$ & $(0.6, 0.2, 0.2)$  \\
Electricity & $26,304$ & $321$ & $168$ & $(0.6, 0.2, 0.2)$ \\
\hline
\end{tabular}
}
\caption{The statistics of datasets.}
\label{tab:exp setting}
\end{table}

\subsection{Synthetic Dataset}
\label{sec:appendix synthetic}
The synthetic dataset used in \Cref{sec:temporal analysis} is a uni-variate time series in the format of
\begin{align}
    \mathcal{X}(t) = A_1\sin{(\omega_1 t)} + A_2\sin{(\omega_2 t + \phi)} + \mathcal{N}(0, \sigma).
\end{align}
The synthetic time-series data is the combination of 3 independent terms.
The first term uses a small $\omega_1 = 5\times 10^{-3}$ to emulate the trend of the time series.
We control the $\omega_2$ in the second term to set the seasonality to different frequencies.
The third term adds a Gaussian noise to the sequence as the residual. 
For low-frequency data, we set $A_1 \sim \operatorname{Uni}(1, 5)$, $A_2\sim \operatorname{Uni}(1, 2)$, $\omega_2=0.04\pi$, and $\sigma=0.3$.
For high-frequency data, we set $A_1=9$, $A_2=8$, $\omega_2=0.1\pi$, $\phi \sim \operatorname{Uni}(0, 10)$, and $\sigma=0.5$.

\subsection{Implementation Details}
To construct our proposed SNNs, we use two Pytorch-based frameworks: SnnTorch \cite{Eshraghian2021TrainingSN} and SpikingJelly \cite{SpikingJelly}.
For all SNNs, we set the time step $T_s = 4$.
For all LIF neurons in SNNs, we set threshold $U_{thr} = 1.0$, decay rate $\beta=0.99$, $\alpha=2$ in surrogate gradient function.

\paragraph{Spike-TCN}
% We use a convolutional neural network with an output channel set to $4$ as the spike encoder.
For spike temporal blocks, we set the kernel size to $3$ (short term) or $16$ (long term), the output channel to $16$, and the downsampling channel to $1$.
The time step of SNNs is set to $4$.
We totally use $3$ blocks to construct our Spike-TCN.

\paragraph{Spike-RNN}
% We use a convolutional neural network with an output channel set to $4$ as the spike encoder.
In the recurrent cells, we have configured two linear layers with input and output dimensions set to $128$.
The time step of SNNs is set to $4$.
This setup is consistent across the GRU cell as well.

\paragraph{iSpikformer}

% We use a convolutional neural network with an output channel set to $4$ as the spike encoder.
We set the number of Spikformer blocks to $2$, and the threshold of spiking neurons in the spiking self-attention (SSA) module as $0.25$.
The time step of SNNs is set to $4$.
Furthermore, we set the dimensional spike-form feature $D$ as $512$, and the hidden feature dimension of the feed-forward layer as $1024$.

\paragraph{Training hyper-parameters}
we set the batch size as $128$ and adopt Adam \cite{kingma2014adam} optimizer with a learning rate of $1\times 10^{-4}$.
We adopt an early stopping strategy with $30$ epochs tolerance.
We run our experiments on $4$ NVIDIA RTX A6000 GPUs.

\subsection{Evaluation metrics}
In order to evaluate our model performances, we employ the Root Relative Squared Error (RSE) and the coefficient of determination (R$^2$) calculated as:
\begin{align}
    \mathrm{RSE}&=\sqrt{\frac{\sum_{m=1}^M||\mathbf{Y}^m-\hat{\mathbf{Y}}^m||^2}{\sum_{m=1}^M||\mathbf{Y}^m-\bar{\mathbf{Y}}||^2}}, \\
    R^2&=\frac1{MCL}\sum_{m=1}^M\sum_{c=1}^C\sum_{l=1}^L\left[1-\frac{(Y^m_{c,l}-\hat{Y}^m_{c,l})^2}{(Y^m_{c,l}-\bar{Y}_{c,l})^2}\right].
\end{align}
In these equations, $M$ represents the size of the test sets, $C$ is the number of channels, and $L$ is the prediction length. $\bar{\mathbf{Y}}$ is the average of $\mathbf{Y}^m$, $Y^m_{c,l}$ denotes the $l$-th future value of the $c$-th variable for the $m$-th sample, $\bar{Y}_{c,l}$ is the average of $Y^m_{c,l}$ across all samples, and $\hat{\mathbf{Y}}^m$ and $\hat{Y}_{c, l}^{m}$ denote the ground truths.

Compared to Mean Squared Error (MSE) or Mean Absolute Error (MAE), these metrics are more robust to the absolute values of the datasets and thus widely used in the time-series forecasting setting.

\section{Theoretical Energy Consumption Calculation}
\label{sec:appendix energy consumption}

According to \citet{horowitz20141} and \citet{yao2022attention}, for SNNs, the theoretical energy consumption of layer $l$ can be calculated as: 

\begin{equation}
\label{formula:SNN Energy}
\begin{aligned}
    Energy(l) = E_{AC}\times SOPs(l)
\end{aligned}
\end{equation}

where SOPs is the number of spike-based accumulate (AC) operations.
For traditional artificial neural networks (ANNs), the theoretical energy consumption required by the layer $b$ can be estimated by 

\begin{equation}
\label{formula:ANN Energy}
\begin{aligned}
    Energy(b) = E_{MAC}\times FLOPs(b)
\end{aligned}
\end{equation}

where FLOPs is the floating point operations of $b$, which is the number of multiply-and-accumulate (MAC) operations.
We assume that the MAC and AC operations are implemented on the 45nm hardware \citep{yao2022attention}, where $E_{MAC} = 4.6pJ$ and $E_{AC} = 0.9pJ$.
Note that $1$J $= 10^{3}$ mJ $=10^{12}$ pJ.
The number of synaptic operations at the layer $l$ of an SNN is estimated as 

\begin{equation}
\label{formula:convert}
\begin{aligned}
    SOPs(l) = T \times \gamma \times FLOPs(l)
\end{aligned}
\end{equation}

where $T$ is the number of times step required in the simulation, $\gamma$ is the firing rate of the input spike train of layer $l$.

We want to emphasize that, although this approach to estimating power consumption has been used in many SNN algorithm papers, it is too simplified to show the real power consumption of SNNs in real-world scenarios.
\citet{yao2022attention} think that although this estimation of energy consumption ignores the hardware implementation basis and the temporal dynamic of spiking neurons, it is still useful for simple analysis and evaluation of algorithm performance and guidance for algorithm design.

Besides, the inference speed of SNNs can not be estimated correctly \textbf{on GPUs} because the values $0$ and $1$ will still be processed to the \textbf{float-32} format.
Once an SNN is well-trained, it can be deployed on neuromorphic hardware for inference, where $0$ and $1$ are treated as spike events.
This enables significantly faster inference speed compared to traditional ANN implementations.
If we just run SNNs on GPUs, then the inference speed of SNNs will always be $T_s$ times that of ANNs, where $T_s$ is the number of time steps of SNNs.

\section{Case Study}\label{app:case study}
In this section, we will show the ground truth and prediction of our SNNs in Metr-la ($L=24$) and Solar ($L=24$)
Before we dive into the case study, it's crucial to emphasize that these two datasets present formidable challenges for prediction. This difficulty stems from the fact that each sample in these datasets includes a significant number of missing values, primarily occurring during nighttime periods when both traffic and solar energy tend to approach zero levels.

\begin{figure}[htp]
\centering
\subfigure[]{
\centering
\includegraphics[width=0.21 \textwidth]{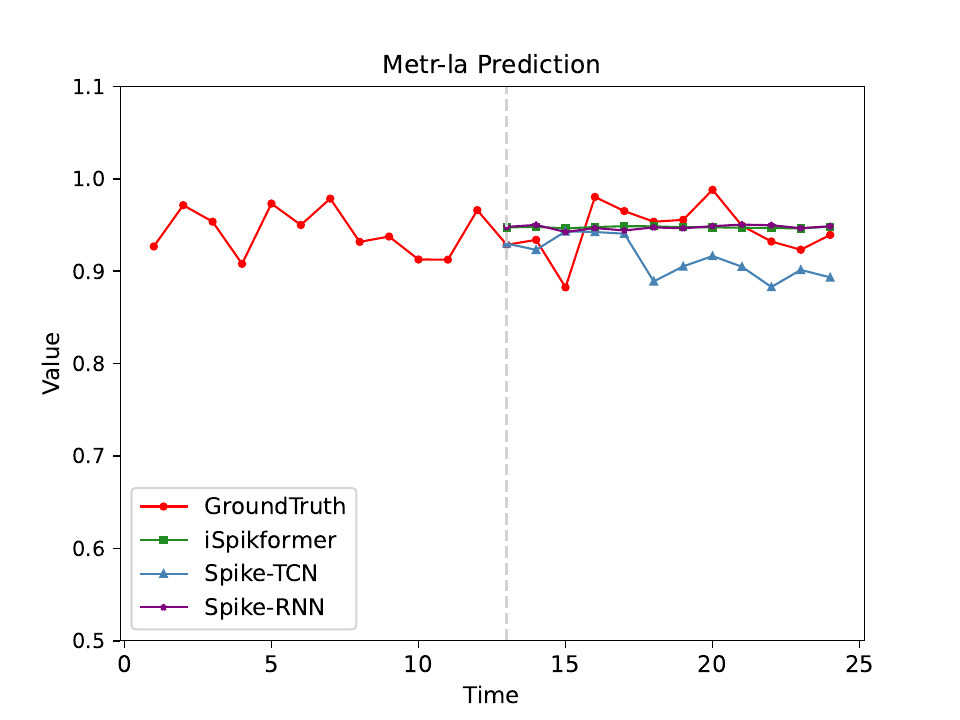}
}
\subfigure[]{
\centering
\includegraphics[width=0.21 \textwidth]{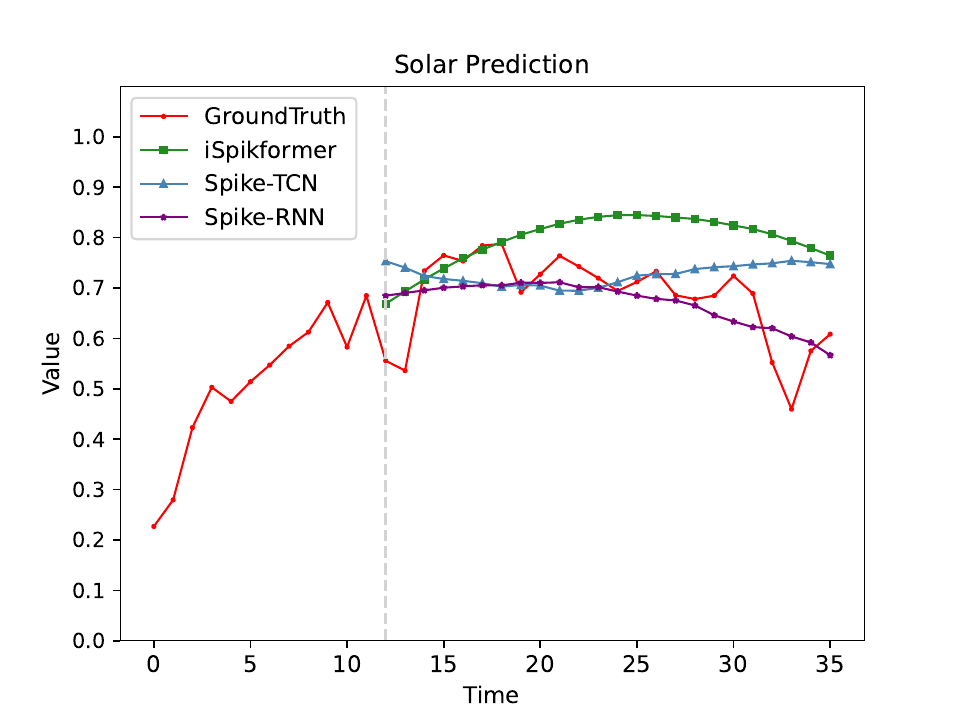}
}
\caption{\label{fig:case study}
Case Study
(a) A prediction slice ($T=12, L=24$) of Spike-TCN, Spike-RNN, and iSpikformer on Metr-la.
(b) A prediction slice ($T=168, L=24$) of Spike-TCN, Spike-RNN, and iSpikformer on Solar.
}
% \vspace{-3mm}
\end{figure}

As depicted in \ref{fig:case study} (a) and (b), all three models exhibit the capability to effectively capture the change trends within the non-missing value portion of the Metr-la dataset. Additionally, these models can roughly simulate the change trends in the Solar dataset as well.
The ability of our proposed SNN model to accomplish this on such a complex dataset underscores its capacity to capture important time series features.
% \begin{table}[h]
% \centering
% \resizebox{\linewidth}{!}{
% \begin{tabular}{|c|c|c|c|c|}
% \hline
% Time Step & Ground Truth & Spike-TCN & Spike-RNN& iSpikformer \\
% \hline
% 1 & 0.9285714 & 0.9299488 & 0.9479422 & 0.9473694 \\
% 2 & 0.93392855 & 0.923069 & 0.95013344 & 0.94799066 \\
% 3 & 0.8825397 & 0.94307935 & 0.9424132 & 0.94628495 \\
% 4 & 0.98035717 & 0.94246405 & 0.9466614 & 0.94779164 \\
% 5 & 0.96507937 & 0.9405746 & 0.9441009 & 0.9486985 \\
% 6 & 0.95357144 & 0.88896894 & 0.9473818 & 0.9485666 \\
% 7 & 0.9555555 & 0.90516984 & 0.9464048 & 0.9475668 \\
% 8 & 0.9880952 & 0.9163733 & 0.94878376 & 0.94746715 \\
% 9 & 0.94920635 & 0.9048208 & 0.95042276 & 0.9470889 \\
% 10 & 0.93214285 & 0.8828873 & 0.9497873 & 0.94664544 \\
% 11 & 0.92321426 & 0.9014037 & 0.94653916 & 0.9463771 \\
% 12 & 0.9392857 & 0.8933017 & 0.94870734 & 0.9481859 \\
% 13 & 0.9392857 & 0.85855126 & 0.95060945 & 0.94911474 \\
% 14 & 0.93214285 & 0.9017634 & 0.95028514 & 0.9488133 \\
% 15 & 0.9539683 & 0.9169718 & 0.95067614 & 0.9481523 \\
% \hline
% 16 & 0.0 & 0.90319294 & 0.94872856 & 0.9480406 \\
% 17 & 0.0 & 0.9184946 & 0.95133096 & 0.94801223 \\
% 18 & 0.0 & 0.92112637 & 0.95224005 & 0.9478716 \\
% 19 & 0.0 & 0.8821734 & 0.9530137 & 0.94767666 \\
% 20 & 0.0 & 0.9004951 & 0.95293254 & 0.9478125 \\
% 21 & 0.0 & 0.88737833 & 0.95163566 & 0.9477163 \\
% 22 & 0.0 & 0.90235174 & 0.9530647 & 0.9480068 \\
% 23 & 0.0 & 0.8910757 & 0.95241064 & 0.947391 \\
% 24 & 0.0 & 0.8872768 & 0.9523708 & 0.94728166 \\
% \hline
% \end{tabular}
% }
% \caption{Ground Truth and Predictions on Metr-la $L=24$}
% \end{table}

\section{Limitations and Future Directions} \label{app:limit}
\subsection{Limitations.}
\paragraph{Spike Decoder.}
The limitations of our work can be summarized as follows.
In the case of SNNs for classification tasks, the use of a fully connected layer for predicting classification results is common.
The classification results can be derived by counting the firing rate of spike trains within each classification category.
Consequently, previous research often omitted the spiking neuron layers added to the projection layer.
However, in regression tasks like time-series forecasting, we rely on the floating-point output of the final projection layer.
In future work, we aim to explore ways to bridge this gap and potentially find methods that allow for a more unified approach in both classification and regression tasks within the SNN framework.

\paragraph{Architecture Designing.}

The study appears as an \textbf{engineering study}, i.e., replacing different parts of the model and evaluating how the model performance will be affected.
In the long run, it would be much more insightful to consider a deeper \textbf{theoretical analysis} to tightly connect the encoder, spiking network model, and decoder together.
Our desired framework is decoupled.   The design of our spike encoders is inspired by the concept of temporal alignment (see \Cref{sec:spike encoder}).
The architecture of our SNNs benefits from ANNs.
Experimental results validate the effectiveness of each component, and good performance is achieved across various backbones (CNN, RNN, Transformers).
Our motivation is to introduce a unified framework for SNNs tailored specifically for time-series forecasting tasks.
What's more, current SNN works on model architecture can be roughly divided into two parts:
(1) designing SNN modules inspired by biological theories and principles;
(2) adapting existing ANN modules to suit the requirements of SNNs.   We agree that while the former is theoretically robust, it poses significant challenges in terms of design and evaluation, whereas the latter offers a more accessible route for SNN development, hence its popularity among researchers.
A fundamental aspect of our contribution lies in bridging the gap between ANNs and SNNs, thereby providing a standardized guideline for the broader community.
Moving forward, we intend to delve deeper into theoretical analyses to elucidate the intricate connections between various SNN modules, building upon the insights garnered from our empirical investigations.

\subsection{Future directions.}
Future research directions encompass:
(1) Developing hardware-friendly algorithms that enable parallel training of TCN-like models without the need to reset $U(t)$ to $0$.
(2) Exploring more promising methods to enhance the utilization of SNNs for capturing \textbf{spatial information} within time-series data, such as spiking graph neural networks.
\end{document}